\documentclass[pdflatex,sn-mathphys-num]{sn-jnl}

\usepackage{graphicx}%
\usepackage{multirow}%
\usepackage{amsmath,amssymb,amsfonts}%
\usepackage{amsthm}%
\usepackage{mathrsfs}%
\usepackage[title]{appendix}%
\usepackage[table]{xcolor}
\usepackage{textcomp}%
\usepackage{manyfoot}%
\usepackage{booktabs}%
\usepackage{algorithm}%
\usepackage{algorithmicx}%
\usepackage{algpseudocode}%
\usepackage{listings}%

\theoremstyle{thmstyleone}%

\theoremstyle{thmstyletwo}%

\theoremstyle{thmstylethree}%

\begin{document}

\title[Reject-to-Remeasure]{Rejections Based on Predictive Uncertainty Enable Reliable Routine Soil Spectroscopy}

\author*[1,2]{\fnm{Jonas} \sur{Schmidinger}}
\email{Jonas.Schmidinger@uni-osnabrueck.de}

\author[2]{\fnm{Robin} \sur{Gebbers}}

\author[3]{\fnm{Marc-Olivier} \sur{Gasser}}

\author[1,2]{\fnm{Viacheslav} \sur{Barkov}}

\author[3]{\fnm{G.Mick} \sur{Wu}}

\author[4]{\fnm{Viacheslav I.} \sur{Adamchuk}}

\affil*[1]{\orgname{Joint Lab Artificial Intelligence and Data Science, Osnabrück University}, \orgaddress{\city{Osnabrück}, \country{Germany}}}

\affil[2]{\orgname{Department of Agromechatronics, Leibniz Institute for Agricultural Engineering and Bioeconomy (ATB)}, \orgaddress{\city{Potsdam}, \country{Germany}}}

\affil[3]{\orgname{Research and Development Institute for the Agri-environment (IRDA)}, \orgaddress{\city{Québec City}, \country{Canada}}}

\affil[4]{\orgname{Department of Bioresource Engineering, McGill University}, \orgaddress{\city{Montréal}, \country{Canada}}}

\abstract{Soil properties relevant to agricultural and environmental applications are conventionally measured using elaborate laboratory methods involving physical and chemical processing. While highly accurate, these conventional methods are costly and time-consuming. In contrast, optical spectroscopy paired with machine learning enables rapid and cost-effective predictions of multiple soil properties. However, spectroscopic modelling is often considered unreliable, as the predictive accuracy varies between soil properties and individual samples. To balance this trade-off between cost and reliability, we introduce reject-to-remeasure: an AI-based measurement framework that combines probabilistic modelling with uncertainty-guided rejection. In this framework, soil samples are first analysed using spectroscopy, after which predictions are rejected if their predictive uncertainty exceeds predefined quality constraints. Rejected samples are subsequently remeasured using conventional laboratory procedures. On a regional visible–near–infrared spectral soil library from Québec, we demonstrate that reject-to-remeasure with modern foundation models (TabPFNv2.5 and TabICLv2) can facilitate the integration of optical spectroscopy into routine laboratory workflows while meeting user-defined accuracy requirements and reducing measurement costs.}


\maketitle

\section{Introduction}\label{sec1}

High-resolution soil maps and frequent soil monitoring are fundamental for precision agriculture. For example, mapping soil nutrient concentrations enables optimisation of fertiliser application \cite{gebbers2010precision}, and monitoring soil organic carbon provides essential information for carbon sequestration \cite{minasny2017soil}. The reliability of soil maps for these purposes ultimately depends on a sufficient number of high-quality reference soil observations \cite{brodsky2013uncertainty,vzivzala2024soil}.

Reference soil observations are typically obtained through laboratory analyses, where soil properties are measured using elaborate procedures involving physical and chemical processing. Although conventional laboratory measurements are relatively accurate, they are also costly and time-consuming. As a result, economic constraints often reduce soil sampling densities and monitoring frequencies, thus preventing optimal adoption of site-specific soil management \cite{rossel2016soil}.

In contrast to conventional laboratory methods, sensor-based approaches such as visible–near–infrared spectroscopy (VNIRS) require little to no sample processing. Hence, VNIRS offers an inexpensive and rapid alternative to acquire soil information \cite{rossel2006determining,nocita2015soil}. For dried soil samples, a spectrometer produces a reflectance spectrum that is strongly influenced by the composition of minerals, and organic materials \cite{clark1990high,ben1995near,rossel2006determining}. Through statistical modelling, this spectrum is used to make quantitative predictions of multiple soil properties \cite{ben1995near,lozano2026accuracies}. Contemporary modelling based on VNIRS increasingly relies on advanced machine learning (ML) techniques \cite{ng2019convolutional,barkov2026modern} trained on spectral libraries \cite{rossel2016global,safanelli2025open}.

Despite cost-efficiency and ongoing advances in modelling, the deployment of VNIRS in soil laboratories remains limited to this day \cite{poppiel2022bridging,sharififar2025navigating}. Moreover, scepticism about the suitability of VNIRS for routine soil analysis was expressed by McBride \cite{mcbride2022estimating} and Baveye \cite{baveye2022vnirs}. The main criticism has been that the quality of spectroscopy-based predictions differs drastically across commonly assessed soil properties. While organic matter and clay fractions can be determined with high accuracy, many essential nutrients do not produce distinct spectral signals at typical concentrations in the soil matrix and are therefore less reliably predicted \cite{safanelli2025open}. For this reason, Baveye \cite{baveye2022vnirs} stated that ``[\ldots] VNIRS should in general be considered fundamentally inadequate as a substitute for traditional, wet-chemistry soil testing methods for many soil chemical properties of interest''.

Although there is an inherent tension between measurement accuracy and measurement cost, we argue that the discussion is too often reduced to a false dilemma, forcing a choice between conventional laboratory analysis and VNIRS. Instead, strategies are needed that explicitly account for predictive uncertainties in modelling. In such an approach, the choice between VNIRS and conventional analysis could be sample- and property-specific, with conventional laboratory analysis reserved for cases in which the spectroscopy-based prediction is expected to be inaccurate. Previous approaches suggested unsupervised outlier detection to flag such samples \cite{poppiel2022bridging,de2023systematic,shepherd2002development}, but predictive uncertainty can be quantified explicitly through probabilistic modelling \cite{gneiting2007probabilistic,schmidinger2023validation}. A probabilistic model not only issues a single prediction value (i.e., point prediction) but also provides a predictive distribution, which represents the model’s belief about the range of plausible values. While probabilistic ML has been explored in soil spectroscopy for providing prediction intervals for end-users (e.g., \cite{padarian2022assessing,wadoux2025uncertainty}), it remains largely unresolved how predictive uncertainty can be operationalised within explicit decision frameworks \cite{wadoux2021ten,lark2022decisions}.

One decision-theoretic way of translating predictive uncertainty into actionable decisions is a predictive model with a rejection option \cite{Chow1970OnOR}. In this approach, a model can abstain from issuing a prediction when the estimated uncertainty is deemed too high given defined thresholds. This idea has increasingly gained attention in medical AI-guided decision frameworks \cite{zhang2023survey,hendrickx2024machine}, where a model may return ``I do not know'' as an output, to prompt further information gathering on a condition or to defer the task to a knowledgeable expert \cite{rodriguez2011spectral,kompa2021second}. Mirroring this logic for soil spectroscopy, an incoming soil sample would first be measured using VNIRS, and an array of soil properties would be immediately predicted from the spectrum. However, depending on sample- and property-specific uncertainty, individual predictions may be rejected, thereby prompting remeasurement with conventional laboratory analysis. We refer to this conceptual framework as reject-to-remeasure. Given that spectroscopic sensing costs only a fraction of conventional analyses \cite{nduwamungu2009opportunities}, reject-to-remeasure could reduce overall analytical costs, even if a considerable proportion of VNIRS predictions is rejected and remeasured by conventional methods. 

For reject-to-remeasure to be effective, the underlying probabilistic model must yield reliable and informative predictive distributions \cite{zhang2023survey}. Reliability means that predicted probabilities match the frequencies observed in independent test data. At the same time, predictive distributions should be as sharp as possible to be informative, since overly wide distributions lead to excessive rejection rates. These two properties are inherently linked, as sharpness should be maximised under the constraint of reliability \cite{gneiting2007probabilistic}. Finally, the model must be discriminative to express varying levels of uncertainty across samples, so that uncertain predictions can be distinguished from confident ones \cite{zhang2023survey}. This is required for achieving conditional reliability, meaning that predicted probabilities match empirical frequencies also under varying feature conditions and across the whole target range \cite{braun2025conditional}. Proper scoring rules, such as the continuous ranked probability score (CRPS), jointly assess reliability, sharpness and discriminative ability \cite{brocker2009reliability}, thereby enabling the optimal selection and tuning of models for reject-to-remeasure.

In this work, we introduce reject-to-remeasure (see Section~\ref{sec2}) to support soil analysis with low-cost sensing approches, such as VNIRS. We evaluate the framework using a VNIRS library from agricultural soils in Québec, Canada (see Section~\ref{sec8.1} for data description). We focus on four agronomically relevant soil properties spanning different levels of predictive difficulty: clay and soil organic matter (SOM), which are typically predicted accurately from VNIRS, as well as extractable potassium (K) and phosphorus (P), which are more challenging since they do not generate distinct spectral signals when measured in a soil matrix. Four additional properties (sand, extractable aluminium (Al), total carbon (TC), and pH) are analysed as extended results. We address two key questions: (1) whether modern probabilistic models (see Section~\ref{sec3}) can reliably reject untrustworthy predictions while maintaining acceptable risk levels (see Section~\ref{sec4}), and (2) whether reject-to-remeasure can reduce laboratory measurement efforts (see Section~\ref{sec5}).

\section{Formalisation of reject-to-remeasure}\label{sec2}

Let $x \in \mathbb{R}^d$ denote a VNIRS reflectance spectrum of an arbitrary soil sample and $Y \in \mathbb{R}$ the associated soil property, treated as a random variable. A probabilistic model $f$ issues, for any $x$, a predictive distribution of $Y \mid x$, from which a point prediction $\hat{y}(x)$ can be derived. 

In typical workflows, $\hat{y}(x)$ is directly returned to the user for subsequent decision-making. Here, we instead introduce a reject-to-remeasure framework, in which a binary decision function $g(x) \in \{\text{Accept}, \text{Reject}\}$ determines whether $\hat{y}(x)$ is accepted or rejected. This decision function, referred to as the rejector \cite{hendrickx2024machine}, rejects predictions with predictive uncertainty deemed too high. In such a case, the sample should be remeasured using conventional laboratory methods.

The rejector is configured by two parameters: (i) an error threshold $\delta(Y)$, specifying the maximum acceptable absolute prediction error, and (ii) a risk level $\alpha \in (0,1)$, specifying the tolerated probability that $\lvert Y - \hat{y}(x) \rvert$ exceeds $\delta(Y)$.

Then the rejector decision rule is defined as:
\begin{equation}
g(x) =
\begin{cases}
\text{Accept}, & \text{if } \Pr\bigl(\lvert Y - \hat{y}(x) \rvert \geq \delta(Y) \mid x \bigr) \leq \alpha, \\
\text{Reject}, & \text{otherwise}.
\end{cases}
\end{equation}

Here, $\Pr(\cdot \mid x)$ refers to the estimated probability under the predictive distribution. For a reliable probabilistic model with independent and identically distributed (IID) data, the rejector provides an explicit error-rate guarantee: among accepted predictions, the long-run frequency of errors exceeding $\delta(Y)$ will be smaller than $\alpha$.

$\delta(Y)$ may be defined as a constant absolute threshold, $\delta(Y) = \delta_0$. Alternatively, it may be defined as a proportional (i.e., relative) threshold, $\delta(Y) = \delta_0 + bY$, allowing the acceptable error to scale with the magnitude of $Y$, where $b$ is the scaling coefficient. The latter is often important in agronomic applications, where large errors at low concentrations are more consequential than similar errors at higher concentrations. See Section~\ref{sec8.5} for justification behind the set threshold parameters used for the different soil properties in this study.

\begin{figure}[htbp]
    \centering
    \includegraphics[width=\linewidth]{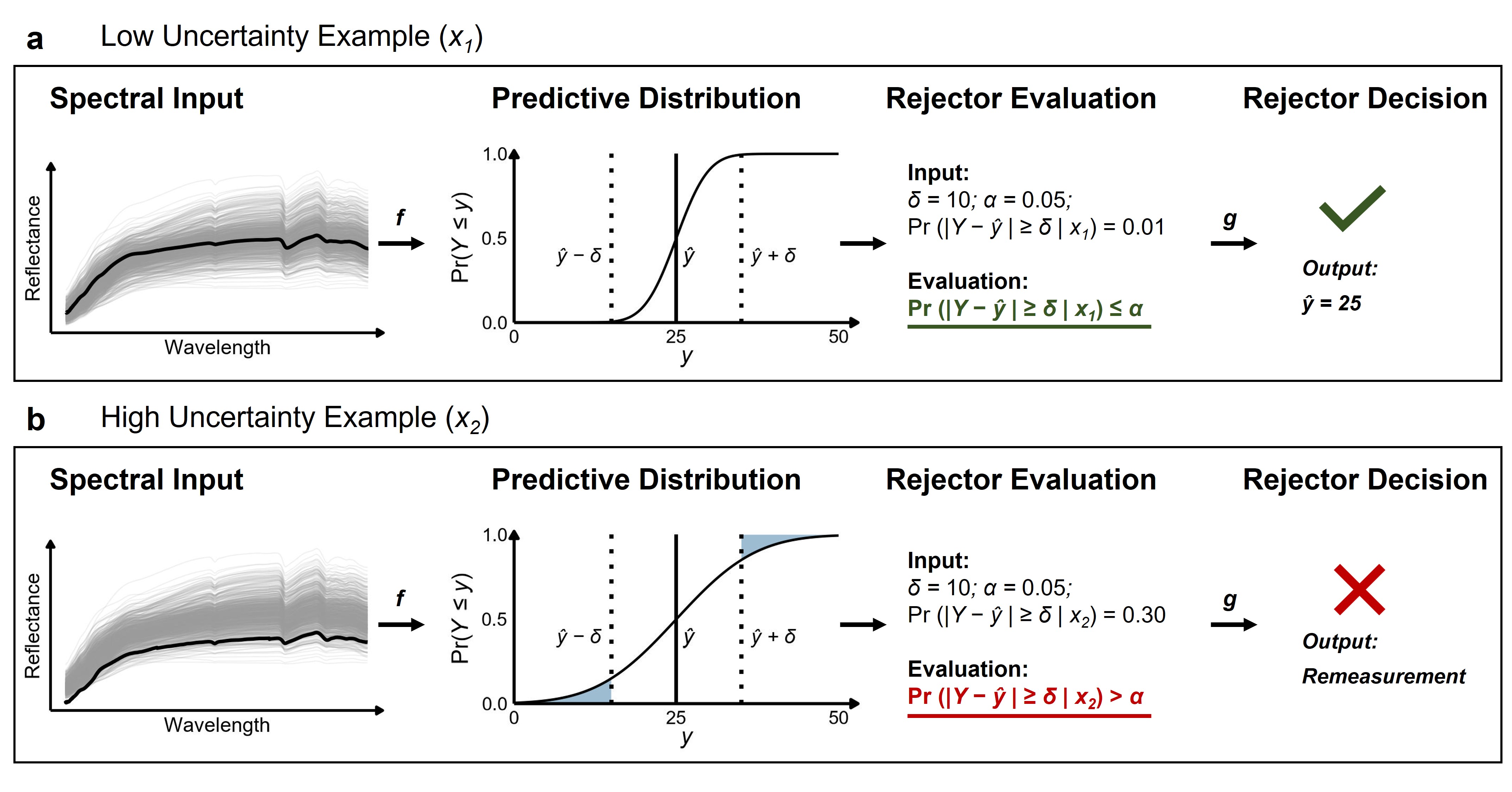}
    \caption{
    \textbf{Workflow of reject-to-remeasure.} The workflow is illustrated for an arbitrary soil property ($y$). The rejector is configured with an absolute error threshold of $\delta = 10$ and a risk tolerance of $5\%$ ($\alpha = 0.05$). The two spectra exemplify a well-represented and less well-represented case. 
    \textbf{(a)} Spectral input ($x_1$) for which the probabilistic model $f$ produces a sharp predictive distribution, here displayed as a cumulative distribution function. The prediction is accepted by the rejector $g$, as the model only issued a $1\%$ probability of exceeding the error threshold. 
    \textbf{(b)} Spectral input ($x_2$) for which the probabilistic model $f$ produces a wide predictive distribution. The prediction is rejected by the rejector $g$, and the sample is remeasured using conventional laboratory analysis because the model issued a $30\%$ probability of exceeding the error threshold.
    }
    \label{fig:Fig1}
\end{figure}

Fig.~\ref{fig:Fig1} illustrates the reject-to-remeasure framework using two spectral examples for predicting an arbitrary soil property. In these examples, the error threshold is set to a constant $\delta(Y) = 10$ (in the respective soil units), and the risk level to $\alpha = 0.05$. This implies that predictions are accepted only if the predicted probability that the absolute error exceeds 10 is below 5\%. Furthermore, we present the framework in an R code-guided tutorial at \href{https://jonasschmidinger.github.io/Reject-to-Remeasure/}{jonasschmidinger.github.io/Reject-to-Remeasure}.

\section{Probabilistic modelling benchmark}\label{sec3}

Because reject-to-remeasure depends critically on the quality of the underlying predictive distributions, we first benchmarked four probabilistic ML models (see Section~\ref{sec8.2} for model implementation) to identify the most suitable models for the subsequent analysis (Section~\ref{sec4} and Section~\ref{sec5}). Table~\ref{tab:Tab1} summarises the predictive performance obtained from the 10-fold cross-validation (CV) (see Section~\ref{sec8.4}) for the four selected soil properties. It shows that the tabular foundation models TabPFN and TabICL consistently outperform Quantile Regression Forest (QRF) and Partial Least Squares Quantile Regression (PLSQR), whose deterministic variants are commonly used in non-probabilistic soil spectroscopy \cite{barra2021soil,gozukara2025prediction}. Table~\ref{tab:TabS1} shows the same results for the additional soil properties.

TabPFN and TabICL generally achieved the highest $R^2$ and lowest mean absolute error (MAE) values. This aligns with recent tabular benchmarks \cite{erickson2025tabarena} and soil-related comparisons \cite{bourriz2025assessing,barkov2026modern}. Beyond their known strong point-prediction abilities, TabPFN and TabICL also exhibited a superior probabilistic performance, as indicated by lower CRPS values. These low CRPS values reflect sharp predictive distributions, given a narrower mean 95\% prediction interval width (MPIW$_{0.95}$), while maintaining reliability, as shown by the empirical quantile coverage probability (QCP). QCP values closely matched the nominal coverage across quantiles (Table~\ref{tab:Tab1} and Fig.~\ref{fig:FigS1}), with deviations typically below 1\%. Nonetheless, both TabPFN and TabICL showed a slight tendency towards overoptimistic behaviour as is common in probabilistic ML, with the extreme quantiles shifted slightly towards the centre of the predictive distribution \cite{schmidinger2023validation}. Even so, the models maintained reasonable conditional coverage (Fig.~\ref{fig:FigS2}), indicating a discriminative ability to capture varying levels of uncertainty across different soil-property ranges.

This model comparison indicates that tabular foundation models may provide the probabilistic capabilities needed for the reject-to-remeasure framework \cite{zhang2023survey}. The performance gap between TabPFN and TabICL was generally small, although TabPFN was slightly stronger overall. At the same time, TabICL, unlike TabPFN, is openly licensed, which might make it more attractive for commercial laboratories. 

For the subsequent rejector evaluation, we restrict our analysis to the best-performing probabilistic model for each soil property, as determined by the lowest CRPS. This corresponds to TabPFN for SOM, P, and K, and to TabICL for clay (outlined by bold letters in Table~\ref{tab:Tab1}).

\begin{table}[ht]
\centering
\setlength{\tabcolsep}{9pt}
\renewcommand{\arraystretch}{1.2}
\caption{
\textbf{Predictive performance of the ML models.} Probabilistic performance was evaluated using CRPS (continuous ranked probability score; lower means better predictive distributions), $\mathrm{MPIW}_{0.95}$ (mean width of the 95\% prediction interval; lower indicates sharper distributions), and $\mathrm{QCP}_{0.025}/\mathrm{QCP}_{0.975}$ (quantile coverage probability; better marginal coverage when closer to 2.5\% and 97.5\%, respectively). Point predictions were evaluated with $R^2$ (higher means better fit) and MAE (mean absolute error; lower means less error). The model with the lowest CRPS (in bold) was selected for the main analysis. See Section~\ref{sec8.3} for details about evaluation metrics.
}
\label{tab:Tab1}
\begin{tabular}{llrrrrrr}
\toprule
Target & Model & CRPS & MPIW$_{0.95}$ & QCP$_{0.025}$ & QCP$_{0.975}$ & $R^2$ & MAE \\
\midrule

\multirow{4}{*}{Clay}
& PLSQR  & 37.61 & 237.44 & 5.30 & 95.08 & 0.87 & 52.37 \\
& QRF    & 35.98 & 260.19 & 3.40 & 97.23 & 0.87 & 49.23 \\
& TabPFN & 23.94 & 168.90 & 2.65 & 97.98 & 0.94 & 32.92 \\
& \textbf{TabICL} & \textbf{23.48} & 169.99 & 2.90 & 98.49 & 0.95 & 32.55 \\

\addlinespace

\multirow{4}{*}{SOM}
& PLSQR  & 6.94 & 43.43 & 5.30 & 93.82 & 0.71 & 10.29 \\
& QRF    & 6.65 & 51.73 & 2.27 & 98.49 & 0.75 & 9.16 \\
& \textbf{TabPFN} & \textbf{4.30} & 29.97 & 2.90 & 96.47 & 0.89 & 5.90 \\
& TabICL & 4.35 & 30.98 & 3.03 & 97.23 & 0.88 & 6.00 \\

\addlinespace

\multirow{4}{*}{K}
& PLSQR  & 32.96 & 234.33 & 6.18 & 95.71 & 0.48 & 49.70 \\
& QRF    & 36.97 & 263.54 & 3.40 & 95.71 & 0.37 & 49.91 \\
& \textbf{TabPFN} & \textbf{31.19} & 222.40 & 3.28 & 96.97 & 0.54 & 45.35 \\
& TabICL & 32.84 & 233.30 & 2.90 & 96.47 & 0.48 & 46.19 \\

\addlinespace

\multirow{4}{*}{P}
& PLSQR  & 27.21 & 176.89 & 7.44 & 93.69 & 0.33 & 42.98 \\
& QRF    & 27.42 & 179.42 & 3.78 & 95.96 & 0.23 & 38.41 \\
& \textbf{TabPFN} & \textbf{23.88} & 158.52 & 3.66 & 95.21 & 0.44 & 34.70 \\
& TabICL & 24.59 & 172.88 & 3.91 & 96.85 & 0.41 & 34.39 \\

\bottomrule
\end{tabular}
\end{table}

\section{Rejector evaluation}\label{sec4}

\begin{figure}[htbp]
    \centering
    \includegraphics[width=\linewidth]{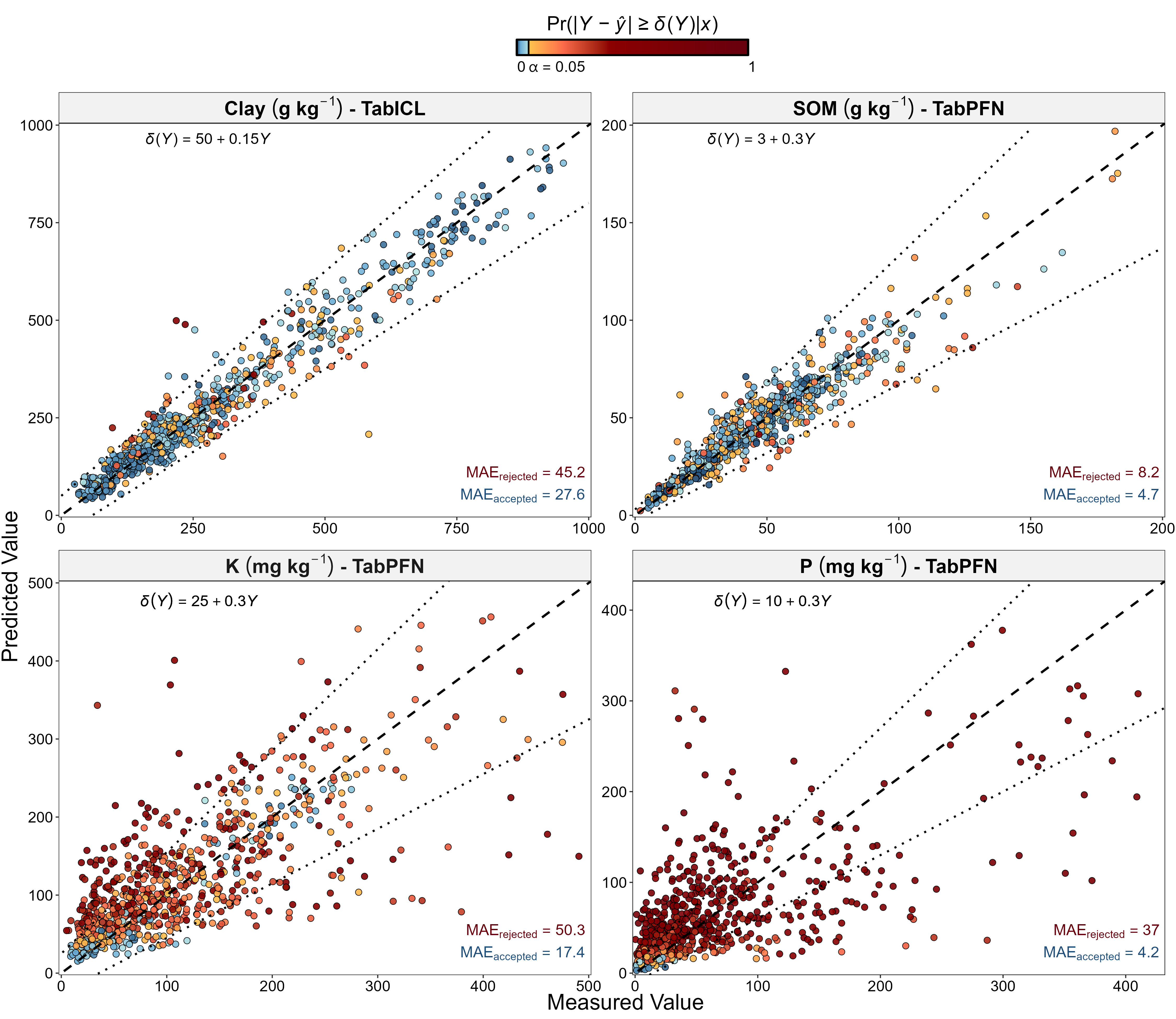}
    \caption{
    \textbf{Predicted versus measured values with uncertainty-based colouring.} Each subplot shows one soil property (Clay, SOM, K, and P), with predicted values plotted against measured values. The dashed line represents the 1:1 predicted–measured relationship, and the dotted lines indicate the property-specific error threshold $\delta(Y)$ (see Section~\ref{sec8.5}). Colours indicate the model’s estimated probability that the absolute prediction error exceeds $\delta(Y)$. Predictions with probabilities below $\alpha = 0.05$ are accepted by the rejector (blue gradient), whereas higher probabilities correspond to rejected predictions (orange–red gradient). $\mathrm{MAE}_{\mathrm{accepted}}$ and $\mathrm{MAE}_{\mathrm{rejected}}$ denote the mean absolute error of the accepted and rejected predictions, respectively. For clarity, five P samples and eight K samples are not displayed because their values fall outside the axis limits.
    }
    \label{fig:Fig2}
\end{figure}

\begin{figure}[htbp]
    \centering
    \includegraphics[width=\linewidth]{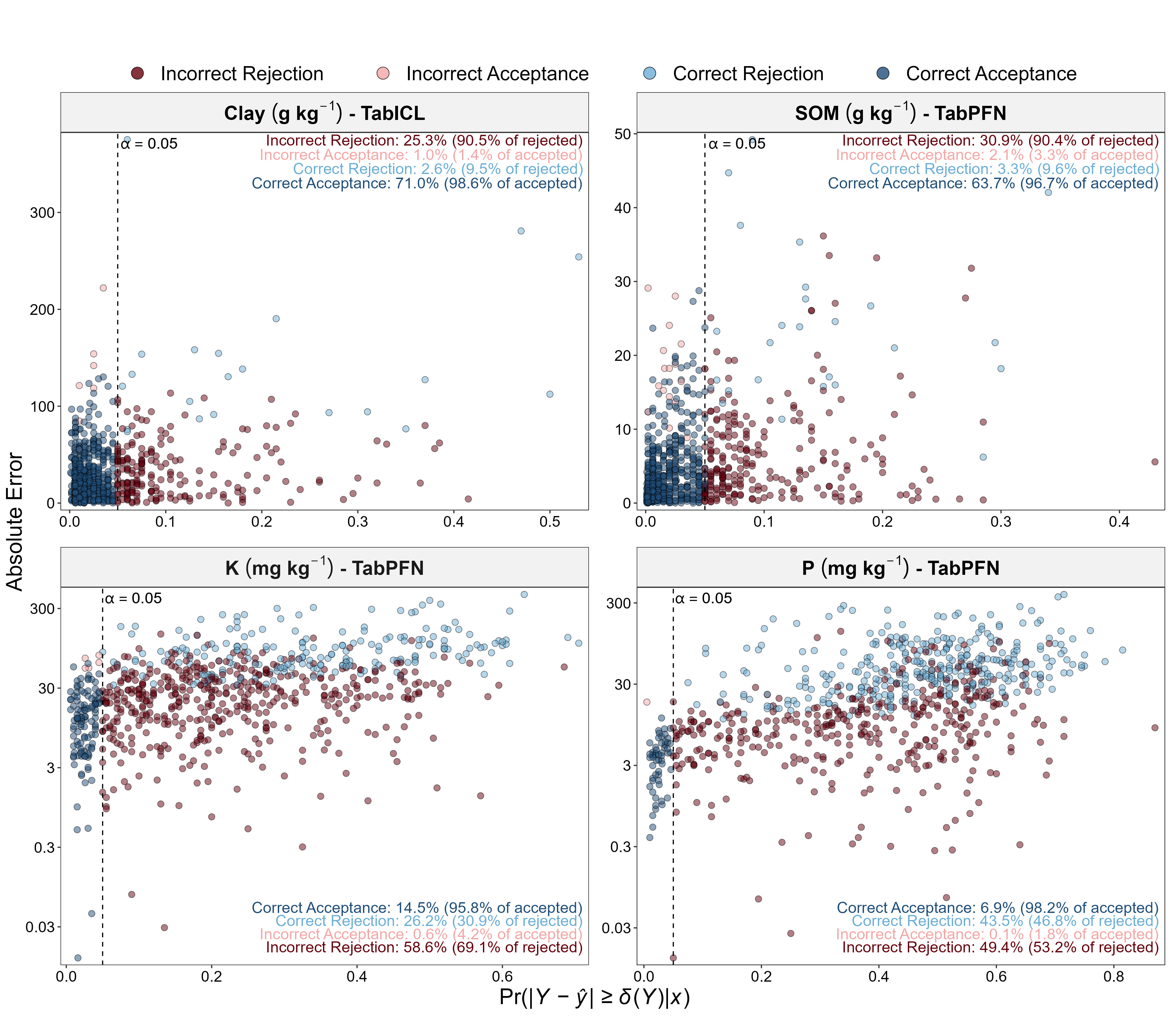}
    \caption{
    \textbf{Rejector decision outcomes.} Each subplot shows one soil property (Clay, SOM, K, and P), with the predicted probability that the absolute error exceeds the threshold $\delta(Y)$, i.e., $\Pr\!\left(\lvert Y - \hat{y}(x) \rvert \geq \delta(Y) \mid x \right)$, plotted against the apparent absolute error. The vertical dashed line indicates the rejector decision boundary corresponding to the risk tolerance $\alpha = 0.05$; predictions to the left are accepted, whereas predictions to the right are rejected. Points are coloured according to the resulting decision. Percentages denote the fraction of samples in each category. Values outside parentheses refer to the percentage relative to all samples, whereas values in parentheses refer to the percentage relative to accepted or rejected predictions. Absolute error of K and P are shown in a log-scale.
    }
    \label{fig:Fig3}
\end{figure}

Across all soil properties, the rejector successfully distinguished between samples with potentially high and low prediction errors, effectively flagging samples prone to large errors (Fig.~\ref{fig:Fig2}, Fig.~\ref{fig:Fig3}, Fig.~\ref{fig:FigS3} and Fig.~\ref{fig:FigS4}). Most importantly, the number of incorrect acceptances---accepted predictions whose error ultimately exceeded the threshold $\delta(Y)$---was reduced to less than 5\% as desired by the set risk level $\alpha = 0.05$. In contrast, the frequency of correct rejections---rejected predictions whose error ultimately exceeded $\delta(Y)$---ranged from 9.5\%--46.8\% (in dark blue, Fig.~\ref{fig:Fig3}). As a direct consequence of this selective rejection of high-error samples, the MAE of accepted predictions was consistently lower than that of rejected predictions (Fig.~\ref{fig:Fig2}).

While $\alpha$ effectively controlled the risk of incorrect acceptances, it is important to note that alternative choices for setting up $\delta(Y)$ may cause the incorrect acceptance rate to slightly exceed 5\% (see Fig.~\ref{fig:FigS5}). We attribute this primarily to the fundamental impossibility of achieving perfect conditional reliability \cite{braun2025conditional} and the minor yet consistent overoptimism noted previously (Fig.~\ref{fig:FigS1}).

The behaviour of the probabilistic models, and thus the rejector, differed considerably between soil properties, depending on the error magnitude and error variance structure (Fig.~\ref{fig:Fig2} and Fig.~\ref{fig:Fig3}). For clay and SOM, the acceptance rates (Fig.~\ref{fig:Fig3}) were relatively high (72\% and 65.8\%, respectively), reflecting that prediction errors were generally small relative to the threshold $\delta(Y)$, such that only a limited fraction of samples required rejection. Predictions were also accepted evenly across their target ranges due to only slight heteroscedasticity, with increased errors occurring in regions with higher error tolerance (Fig.~\ref{fig:Fig2}). In contrast, acceptance rates for K and P were substantially lower (15.1\% and 7.1\%, respectively). In other words, the model was unable to generate confident predictions for the majority of samples. Due to strongly increasing heteroscedastic error patterns, acceptances were also mostly clustered in lower value ranges. In the case of P, all predictions beyond $25~\mathrm{mg\,kg^{-1}}$ were simply rejected. The extended analysis shows further diverse error patterns across different soil properties (Fig.~\ref{fig:FigS3}), including decreasing error variance (for pH) and non-monotonic variance structures (for sand).

Referring back to the critiques raised by McBride \cite{mcbride2022estimating} and Baveye \cite{baveye2022vnirs}, the results for K and P (Fig.~\ref{fig:Fig2}) support the notion that ``VNIRS is fundamentally ill-suited for the measurement, in individual soil samples, of a whole slew of [\ldots] soil chemical parameters'' \cite{baveye2022vnirs}. This is reflected in the high rejection rates (Fig.~\ref{fig:Fig3}), implying that conventional soil nutrient determination cannot be substituted by VNIRS in the foreseeable future. On the other hand, the high acceptance rates observed for clay and SOM indicate substantial potential to partially replace costly conventional analyses with VNIRS. This property-specific performance has been widely recognised; however, rather than treating these differences as generalisations at the property level, reject-to-remeasure formalises them as conditional, sample-specific decisions under explicit accuracy constraints.

We argue that reject-to-remeasure could be applied to any other soil sensing technique, such as mid-infrared spectroscopy, X-ray fluorescence spectroscopy, laser-induced breakdown spectroscopy, and gamma-ray spectroscopy. While these techniques are not yet as widely used as VNIRS, some of them directly estimate total elemental concentrations \cite{lozano2026accuracies}. Fusing VNIRS data with these sensing techniques could increase prediction accuracy \cite{tavares2021multi} and, therefore, raise acceptance rates even for nutrient predictions.

\section{Economic implications}\label{sec5}

\begin{figure}[htbp]
    \centering
    \includegraphics[width=\linewidth]{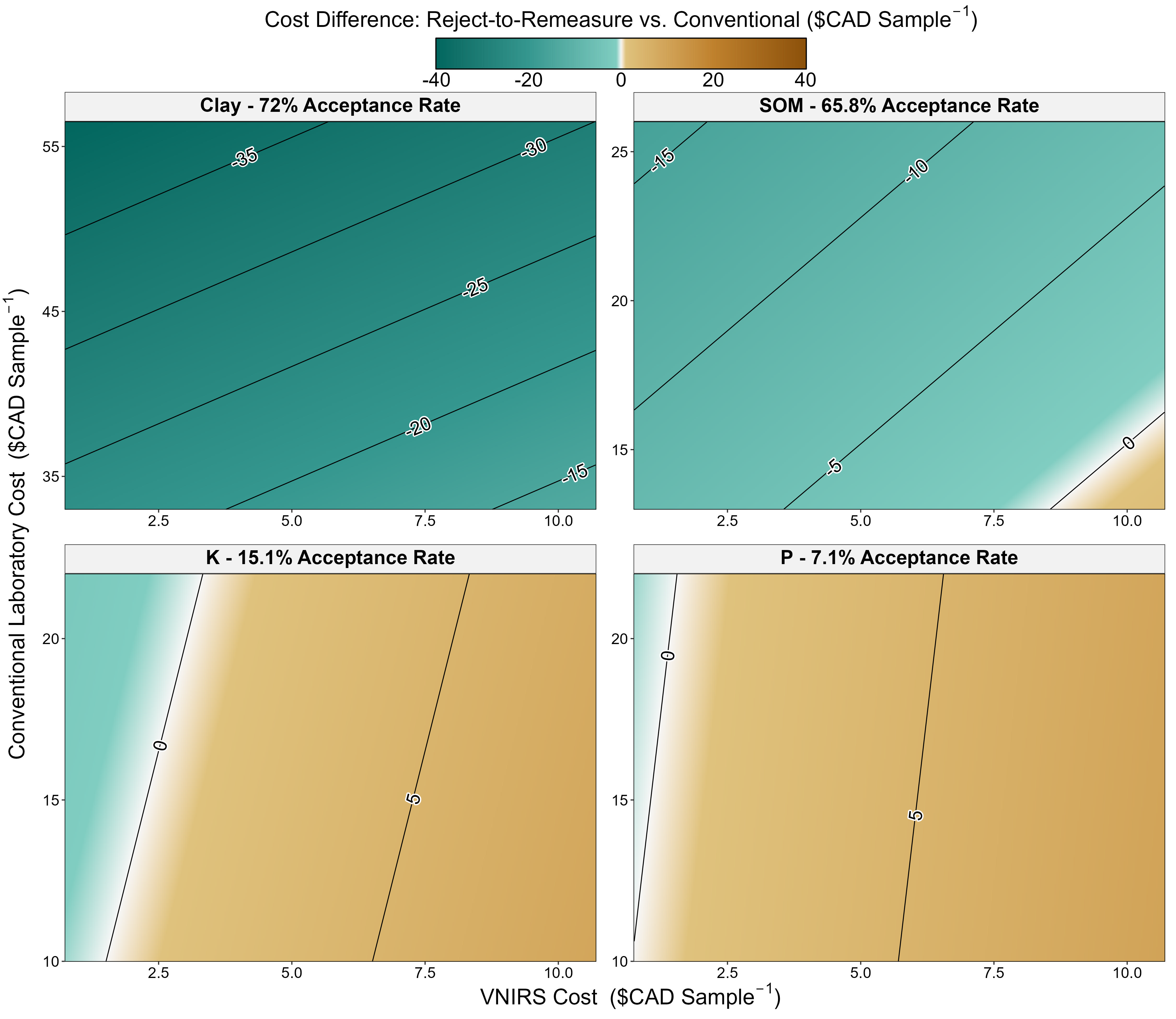}
    \caption{
    \textbf{Cost difference between VNIRS reject-to-remeasure and conventional laboratory testing.} Each subplot shows the cost difference in analysing a soil property (Clay, SOM, K, and P) when using the VNIRS-based reject-to-remeasure strategy compared to pure conventional laboratory analysis. Plausible ranges of measurement costs for conventional analysis and VNIRS were evaluated (see Section~\ref{sec8.6}). Negative values (blue-green) indicate cost savings with reject-to-remeasure as compared to the conventional analysis, whereas positive values (brown) indicate increased measurement costs. Contour lines denote equal levels of cost savings. The plot assumes the acceptance rates obtained in the previous rejector evaluation (see Section~\ref{sec8.5}).
    }
    \label{fig:Fig4}
\end{figure}

Having established that reject-to-remeasure achieves reliable error control, we now evaluate whether this translates into economic benefits for commercial laboratories. In an illustrative economic analysis, we considered measurement cost ranges for conventional soil analyses comparable to those of the IRDA research institute, which analysed the soil samples of this study (see Section~\ref{sec8.1}). We further considered VNIRS measurement costs of 1--10~\$CAD (see Section~\ref{sec8.6} for details). Lastly, we assumed the same threshold constraints for the rejector as in the previous sections. 

For clay and SOM, the results indicate a clear potential for reducing measurement costs using reject-to-remeasure based on VNIRS (Fig.~\ref{fig:Fig4}). This cost reduction resulted primarily from the fact that only a fraction of SOM and clay predictions required remeasurement, due to their high acceptance rates (Fig.~\ref{fig:Fig3}). In addition, there is a large price gap between inexpensive VNIRS measurements and the more laborious particle size determination based on time-consuming sedimentation \cite{nduwamungu2009opportunities}.

In contrast, reject-to-remeasure based on VNIRS may not be economically viable for the determination of K or P. Soil nutrient determination through the Mehlich-3 extraction \cite{mehlich1984mehlich} is already relatively inexpensive, thereby narrowing the cost advantage of VNIRS for these properties. In addition, most K and P predictions were rejected, meaning that a large proportion of samples would need to be remeasured. As a result, many VNIRS scans become effectively redundant and thus increase overall costs.

One of the main advantages of VNIRS is its ability to jointly predict multiple soil properties from a single scan \cite{ben1995near,lozano2026accuracies}. This creates additional potential for cost reduction, as the cost of a VNIRS scan can be distributed across multiple predicted properties, whereas conventional analyses typically require separate laboratory procedures for each property. Consequently, high rejection rates for individual soil nutrients are not necessarily problematic, as the same VNIRS measurement may still yield acceptable predictions for other properties. To avoid introducing additional assumptions, we restricted the present cost comparison to a single property analysis. This makes our estimates conservative, and real operational deployment of reject-to-remeasure may be even more economically attractive. However, a full multi-property cost analysis is beyond the scope of this study and is left for future work.

The cost ratio between VNIRS and conventional laboratory methods varies geographically due to factors such as labour costs and infrastructure \cite{nduwamungu2009opportunities}. Nonetheless, this illustrative economic analysis for a case study in Québec demonstrates that the reject-to-remeasure strategy can be economically viable (Fig.~\ref{fig:Fig4} and Fig.~\ref{fig:FigS6}) while mostly maintaining predefined quality constraints (see Section~\ref{sec4}). Importantly, this approach acknowledges the fundamental limitations of VNIRS highlighted in previous critiques \cite{mcbride2022estimating,baveye2022vnirs}, but provides a framework to navigate these limitations in operational settings. Therefore, reject-to-remeasure represents a concrete step towards the operational integration of soil spectroscopy into routine agricultural and environmental workflows.

\section{Further research and limitations}\label{sec6}

\begin{figure}[htbp]
    \centering
    \includegraphics[width=\linewidth]{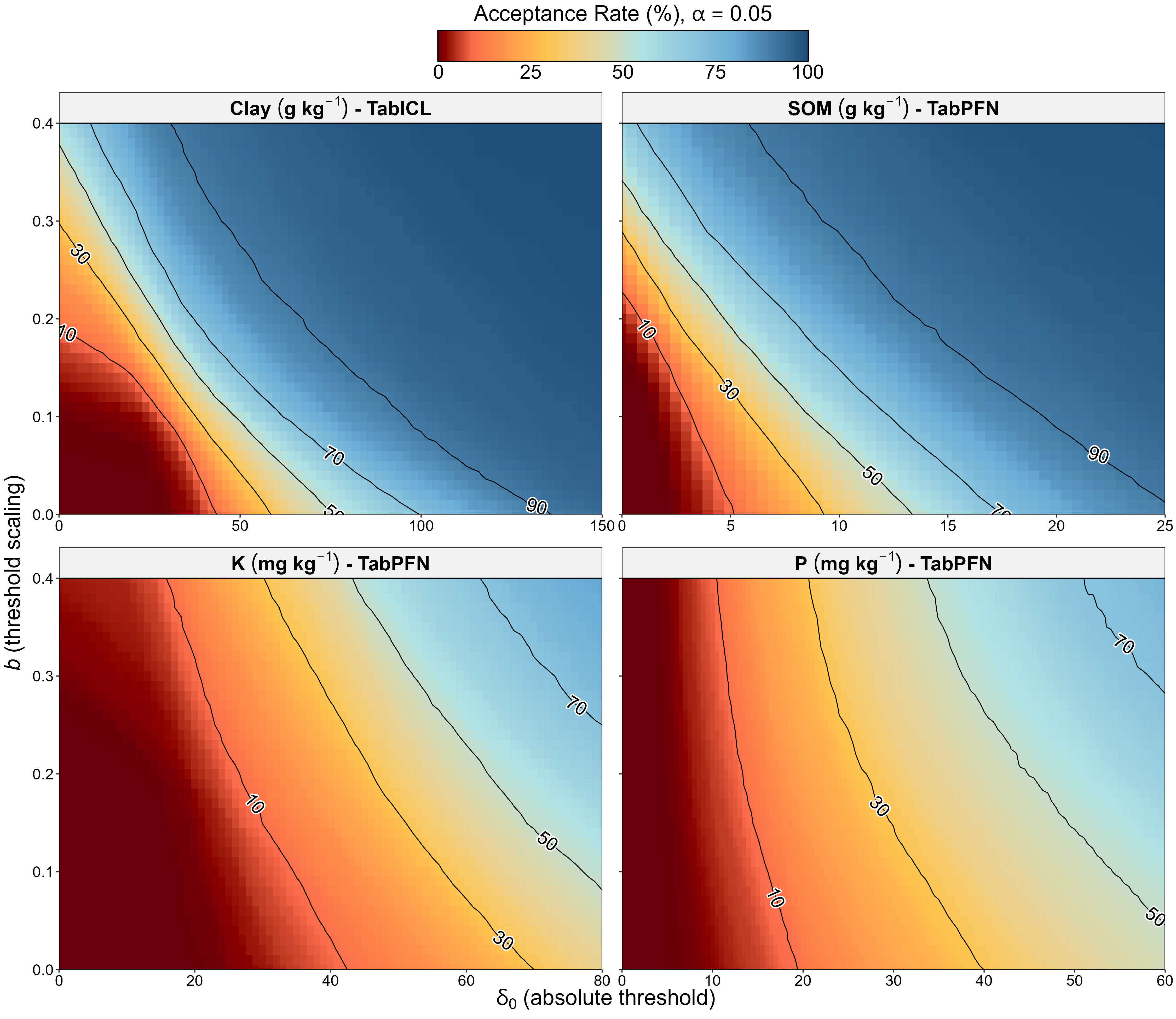}
    \caption{
    \textbf{Acceptance rates relative to the error threshold parameters.} Each subplot shows the acceptance rates for a soil property (Clay, SOM, K, and P) as a function of the parameters used in the error threshold $\delta(Y) = \delta_0 + bY$, where $\delta_0$ is the absolute threshold component and $b$ is the scaling coefficient. With higher error tolerances, acceptance rates increase (blue), whereas overly strict quality constraints reduce the acceptance rate to near zero (red). Contour lines denote equal acceptance rates.
    }
    \label{fig:Fig5}
\end{figure}

Reject-to-remeasure naturally enables the use of task-specific error constraints. For clarity, however, we first considered a single plausible threshold $\delta(Y)$ per soil property (Section~\ref{sec8.5}) for the main analysis. Depending on the specific purpose, the set threshold constraints may be considered either too conservative or too lenient. Therefore, Fig.~\ref{fig:Fig5} and Fig.~\ref{fig:FigS7} illustrate how the acceptance rate changes depending on other plausible ranges of $b$ and $\delta_0$ under the threshold function $\delta(Y) = \delta_0 + bY$. They show that the acceptance rate can rapidly drop to near zero or rise to 100\%, depending on the chosen parameters. This has different implications for the economic viability of the reject-to-remeasure framework. For example, requiring clay predictions to match the precision of conventional particle size analysis of $\delta(Y) = 30~\mathrm{g\,kg^{-1}}$ would result in all predictions being rejected, thereby eliminating the practical utility of VNIRS for clay estimation. In contrast, a more lenient threshold of $\delta(Y) = 40\text{--}50~\mathrm{mg\,kg^{-1}}$ for P would make even nutrient determination with VNIRS viable. However, error propagation analysis \cite{heuvelink1998error,ellinger2019error,sanderman2025application} would be required to determine optimal task-specific constraints. 

Our analysis was conducted on a regional IID dataset, meaning that training and test observations are drawn from the same geographic region, here agricultural soils in Québec. This is a representative scenario for most soil laboratories which usually operate on a restricted regional level. However, when developing a global soil spectroscopy model \cite{rossel2016global,safanelli2025open}, out-of-distribution predictions remain a major challenge \cite{kock2024development,huang2025using}. Testing the utility of reject-to-remeasure for such tasks requires further evaluation of how well foundation models based on in-context learning can adapt and appropriately increase their predictive uncertainty when confronted with soils poorly represented in the training data \cite{huang2025using}. In principle, the reject-to-remeasure framework, combined with retraining on remeasured samples, could enable self-improving AI, as newly analysed samples would gradually expand the representation of previously underrepresented soil types.

\section{Conclusions}\label{sec7}

All too often, soil spectroscopy is presented as a substitute for conventional laboratory analysis, creating a false dilemma between cost efficiency and output reliability. In contrast, the proposed reject-to-remeasure framework treats spectroscopy as the initial measurement method, while reverting to conventional laboratory analysis when the predictive uncertainty exceeds an acceptable level.

Our results, based on a regional VNIRS library, show that reject-to-remeasure can be a viable measurement strategy. By leveraging the strong probabilistic capabilities of modern tabular foundation models, we demonstrate a clear potential to reduce measurement costs while ensuring that accepted predictions meet predefined quality constraints. This provides a pathway towards operational deployment of soil spectroscopy for routine agricultural and environmental applications through AI-guided decision making.

Future work should evaluate reject-to-remeasure for out-of-distribution prediction tasks, where the framework may additionally enable self-improvement through iterative retraining on rejected and subsequently remeasured samples. Beyond soil spectroscopy, the concept could also serve as a general blueprint for other chemometric applications involving different sensing technologies.

\section{Methods}\label{sec8}

\subsection{Data}\label{sec8.1}

\begin{figure}[htbp]
    \centering
    \includegraphics[width=0.8\linewidth]{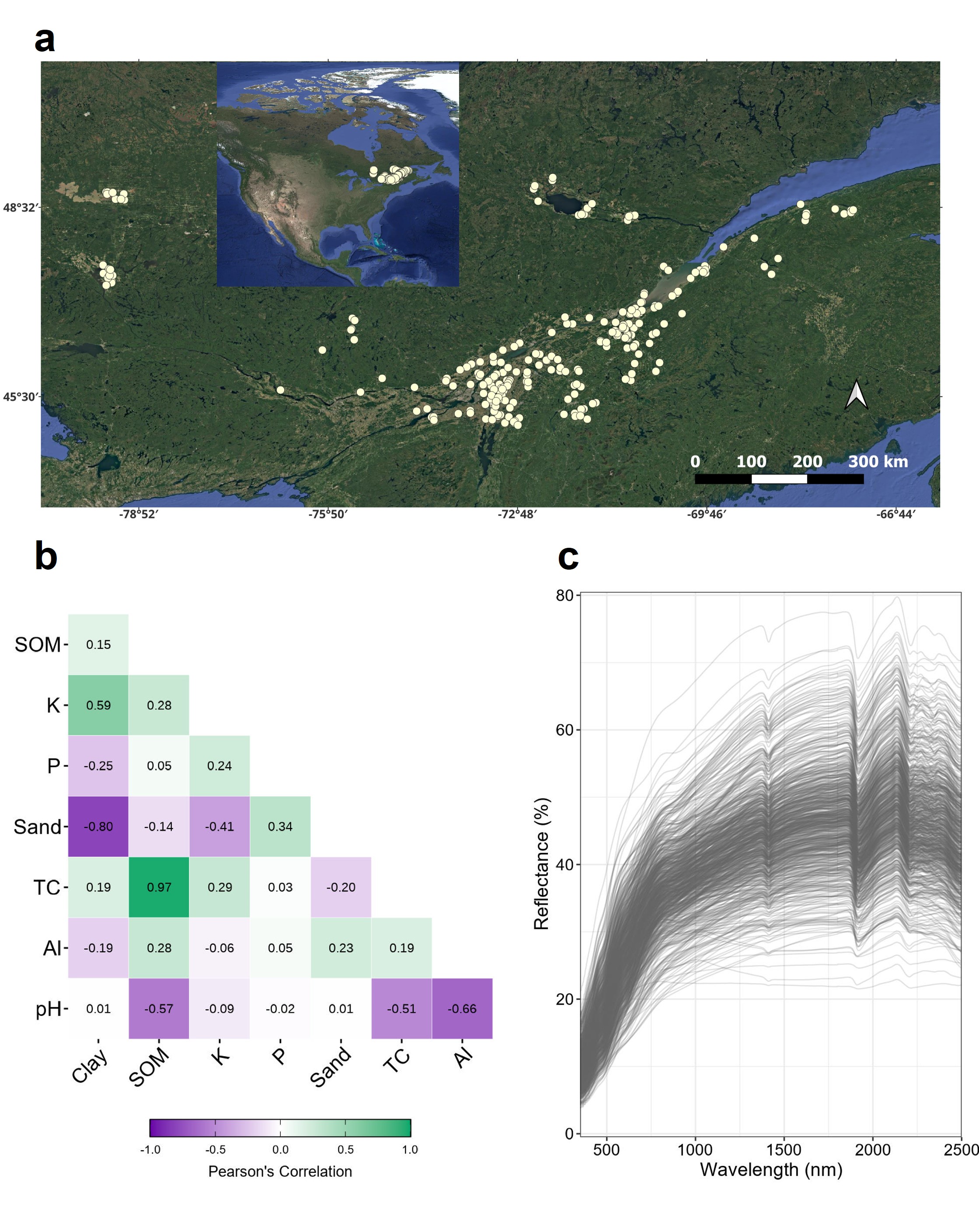}
    \caption{
    \textbf{Dataset overview.} \textbf{(a)} Sampling sites across the agricultural land of Québec (WGS 84). \textbf{(b)} Correlation between the evaluated soil properties, expressed as Pearson’s correlation coefficients. \textbf{(c)} Raw VNIRS spectra.
    }
    \label{fig:Fig6}
\end{figure}

The dataset consists of soil reference samples collected at 272 sites across agricultural land in Québec, Canada (see Fig.~\ref{fig:Fig6}a). At each site, samples were obtained from two to three soil horizons: the first 0--10~cm of topsoil (Ap1 horizon), the second part of the topsoil if deeper than 17~cm (Ap2 horizon), and the first 15~cm of the subsoil (B horizon), resulting in a total of 793 analysed samples. Sampling and reference analyses were conducted from 2018 to 2022 as part of a province-wide soil health assessment by IRDA \cite{gasser2023rapport}, using a stratified random sampling design that represents the diversity of soil series across Québec. Soil series reflect differences in parent material and soil texture.

All reference analyses were performed on air-dried, crushed, and 2~mm-sieved soil samples. Detailed analytical protocols are described in \cite{gasser2023rapport}. In summary, particle size distribution (clay and sand) was determined using a modified Bouyoucos hydrometer method, SOM by loss-on-ignition, and TC by dry combustion via CO$_2$ release. Soil pH was measured in the Shoemaker--McLean--Pratt buffer solution, and extractable Al, K, and P were determined using Mehlich-3 extraction with inductively coupled plasma spectrometry. The intercorrelations among these soil properties are displayed in Fig.~\ref{fig:Fig6}b.

VNIRS scanning was conducted on the same pre-treated (i.e., dried, crused and sieved) samples between November and December 2025 using an ASD\textsuperscript{\textregistered} LabSpec4 Standard-Res (Malvern Panalytical Ltd., USA) equipped with a Muglight accessory. It covers the range of 350--2{,}500~nm with a spectral resolution of 3--10~nm. The associated device software returns a reflectance spectrum interpolated to 1~nm resolution. These raw spectra are displayed in Fig.~\ref{fig:Fig6}c.

\subsection{Models}\label{sec8.2}

Four predictive models with probabilistic capabilities were evaluated: TabPFN, TabICL, PLSQR, and QRF. For all models, predictive distributions were derived from predicted quantiles by constructing a cumulative distribution function as a step function over predefined quantile levels $\tau_i$, where $\tau_i \in (0.001, 0.005, 0.01, \ldots, 0.995, 0.999)$.

TabPFN (in full, tabular prior-data fitted network) is a transformer-based foundation model for tabular data that leverages in-context learning \cite{hollmann2025accurate}. The TabPFNv2.5 regressor used in this study was trained exclusively on synthetic datasets \cite{grinsztajn2025tabpfn}. It approximates Bayesian inference by producing (posterior) predictive distributions, where the prior is defined by the synthetic training data. The model is trained by minimising the cross-entropy loss over predictive distributions, with regression formulated through the discretisation of the target space. Using the TabPFNRegressor from the \texttt{tabpfn} Python package, we set the number of base models in the ensemble (\texttt{n\_estimators}) to 4 and tuned the \texttt{softmax\_temperature} and \texttt{average\_logits}, which control the sharpness of the predictive distribution (Table~\ref{tab:Tab2}).

TabICL (TabICLv2 variant) is similarly a tabular foundation model trained on synthetic datasets, leveraging in-context learning \cite{qu2026tabiclv2}. While sharing the same general approach as TabPFN, TabICL incorporates architectural differences in attention, context scaling, and feature embedding strategies. For probabilistic predictions, rather than discretising the target space as in TabPFN, TabICL employs pinball loss as in quantile regression \cite{koenker1978regression}. TabICL is openly licensed, which may be particularly relevant for operational deployment in soil laboratories. As TabICL does not provide a built-in mechanism to adjust distributional sharpness, we implemented a rescaling function (\texttt{s\_rescaler}) applied to the cumulative distribution function $F$, defined as $F(y) = \Pr(Y \leq y \mid x)$:
\begin{equation}
F'(y) = \frac{F(y)^s}{F(y)^s + (1 - F(y))^s},
\end{equation}
where $F'$ denotes the rescaled cumulative distribution function and $s$ the tunable rescaling parameter (Table~\ref{tab:Tab2}).

Partial least squares regression \cite{wold1984collinearity} remains the most widespread model in soil spectroscopy \cite{barra2021soil,gozukara2025prediction}. It naturally handles the high dimensionality of spectral data through supervised dimensionality reduction, whereas other ML techniques often struggle with such feature representations \cite{schmidinger2025limesoda}. In our PLSQR variant, we used the pinball loss to approximate a predictive distribution. This was implemented by combining \texttt{PLSRegression} and \texttt{QuantileRegressor} from \texttt{scikit-learn} \cite{pedregosa2011scikit} in Python. The number of partial least squares components (\texttt{n\_components}) and the rescaling parameter (\texttt{s\_rescaler}), analogous to TabICL, were tuned (Table~\ref{tab:Tab2}).

QRF \cite{meinshausen2006quantile} extends random forests to produce predictive distributions by aggregating the empirical distribution of observations in the terminal nodes across trees. QRF has a long history in soil modelling \cite{vaysse2017using,szatmari2019spatio} and has been more recently used for soil spectroscopy \cite{wadoux2025uncertainty}. We used the \texttt{RandomForestQuantileRegressor} from the \texttt{quantile\_forest} Python package with 500 trees (\texttt{n\_estimators}). The number of features considered at each split (\texttt{max\_features}) and the rescaling parameter (\texttt{s\_rescaler}) were tuned (Table~\ref{tab:Tab2}).

\subsection{Evaluation metrics}\label{sec8.3}
CRPS was used to rank probabilistic models through a single summary metric that jointly reflects the reliability, sharpness, and discriminative ability of predictive distributions. It was calculated using its energy score representation \cite{gneiting2007probabilistic}:
\begin{equation}
\mathrm{CRPS} = \frac{1}{n} \sum_{i=1}^{n} \left( 
\mathbb{E}_{Y \sim F(\cdot \mid x_i)} \lvert Y - y_i \rvert 
- \frac{1}{2} \mathbb{E}_{Y, Y' \sim F(\cdot \mid x_i)} \lvert Y - Y' \rvert 
\right),
\end{equation}
where $Y$ and $Y'$ are independent draws from the predictive cumulative distribution function $F(\cdot \mid x_i)$, and $n$ denotes the number of evaluated observations.

QCP at nominal quantile level $\tau$ quantifies reliability and, when evaluated across multiple $\tau$, characterises the reliability of probabilistic predictions:
\begin{equation}
\mathrm{QCP}_{\tau} = \frac{1}{n} \sum_{i=1}^{n} \mathbb{I}\bigl(y_i \leq q_{\tau}(x_i)\bigr),
\end{equation}
where $\mathbb{I}$ is the indicator function, evaluating whether the true value $y_i$ is covered by the predicted quantile $q_{\tau}(x_i)$. QCP is related to the more commonly used prediction interval coverage probability, but it can reveal one-sided bias in quantile shifts \cite{schmidinger2023validation}.

The MPIW at nominal level $1 - \gamma$ quantifies the sharpness of a prediction interval and thus indicates distributional sharpness \cite{schmidinger2023validation}:
\begin{equation}
\mathrm{MPIW}_{1 - \gamma} = \frac{1}{n} \sum_{i=1}^{n} \left( 
q_{1 - \gamma/2}(x_i) - q_{\gamma/2}(x_i) 
\right),
\end{equation}
where $q_{\gamma/2}(x_i)$ and $q_{1 - \gamma/2}(x_i)$ refer to the lower and upper quantiles of the central $1 - \gamma$ prediction interval, respectively.

Point prediction performance was evaluated using $R^2$ and MAE, representing relative and absolute performance metrics, respectively:
\begin{equation}
R^2 = 1 - \frac{\sum_{i=1}^{n} (y_i - \hat{y}(x_i))^2}{\sum_{i=1}^{n} (y_i - \bar{y})^2},
\end{equation}

\begin{equation}
\mathrm{MAE} = \frac{1}{n} \sum_{i=1}^{n} \lvert y_i - \hat{y}(x_i) \rvert,
\end{equation}
where $\bar{y}$ refers to the mean of evaluated observations.

To evaluate the validity of reject-to-remeasure, we categorised predictions into four outcomes based on the rejector decision $g(x_i) \in \{\mathrm{Accept}, \mathrm{Reject}\}$ and the error constraint $\delta(Y)$:
\begin{align}
\text{Correct Acceptance} &= \{ i : g(x_i)=\mathrm{Accept},\; |y_i - \hat{y}(x_i)| < \delta(y_i) \}, \\
\text{Incorrect Acceptance} &= \{ i : g(x_i)=\mathrm{Accept},\; |y_i - \hat{y}(x_i)| \geq \delta(y_i) \}, \\
\text{Correct Rejection} &= \{ i : g(x_i)=\mathrm{Reject},\; |y_i - \hat{y}(x_i)| \geq \delta(y_i) \}, \\
\text{Incorrect Rejection} &= \{ i : g(x_i)=\mathrm{Reject},\; |y_i - \hat{y}(x_i)| < \delta(y_i) \}.
\end{align}

\subsection{Modelling pipeline}\label{sec8.4}

Modelling was performed in Python, whereas modelling results were analysed in R. The complete source code is available at \href{https://github.com/JonasSchmidinger/Reject-to-Remeasure}{github.com/JonasSchmidinger/Reject-to-Remeasure}.

Model performance was evaluated using 10-fold CV. To prevent data leakage arising from depth-wise autocorrelation \cite{john2025problematic}, sampling sites were randomly assigned to folds, and all observations from the depth profile at that sampling site were kept within the same fold. In other words, random CV was performed based on grouped samples at the sampling-site level. Hyperparameter tuning and spectral preprocessing were optimised via grid search in a nested 10-fold CV. For each model, 960 parameter configurations were evaluated per inner-loop iteration (Table~\ref{tab:Tab2}).

Spectral preprocessing was conducted in Python with functions that approximate those of \texttt{prospectr} \cite{stevens2014introduction}. This included standard normal variate (\texttt{snv}), downsampling of the spectra by averaging adjacent wavelengths into bins (\texttt{bin\_size}), and Savitzky--Golay filtering (\texttt{sg}). For Savitzky--Golay filtering (\texttt{if sg = True}), we searched different window lengths (\texttt{window\_length}), polynomial orders (\texttt{polyorder}), and derivatives (\texttt{deriv}). The full set of preprocessing parameters and model hyperparameters is provided in Table~\ref{tab:Tab2}.

\begin{table}[ht]
\centering
\setlength{\tabcolsep}{10pt}
\renewcommand{\arraystretch}{1.2}
\caption{Tuning grid for hyperparameters and spectral preprocessing parameters in the nested 10-fold CV. Each model was evaluated based on 960 parameter configurations. }
\label{tab:Tab2}
\begin{tabular}{lll}
\toprule
 & \textbf{Parameter} & \textbf{Search space} \\
\midrule

\multirow{6}{*}{\parbox{3.5cm}{Spectral preprocessing \\ (all models)}} 
& \texttt{snv} & (True, False) \\
& \texttt{bin\_size} & (5, 10, 15, 20) \\
& \texttt{sg} & (True, False) \\
& \texttt{polyorder (if sg = True)} & (1, 2) \\
& \texttt{deriv (if sg = True)} & (1, 2) \\
& \texttt{window\_length (if sg = True)} & (3, 7, 13) \\

\midrule
\multirow{2}{*}{TabPFN}
& \texttt{softmax\_temperature} & (0.9, 1, 1.1, 1.2, 1.3, 1.4) \\
& \texttt{average\_logits} & (True, False) \\

\midrule
\multirow{1}{*}{TabICL}
& \texttt{s\_rescaler} & (0.8, 0.85, 0.9, [\ldots], 1.35) \\

\midrule
\multirow{2}{*}{PLSQR}
& \texttt{n\_components} & (5, 10, 25) \\
& \texttt{s\_rescaler} & (0.9, 1, 1.1, 1.3) \\

\midrule
\multirow{2}{*}{QRF}
& \texttt{max\_features} & (0.35, 0.5, 0.65) \\
& \texttt{s\_rescaler} & (0.9, 1, 1.1, 1.3) \\

\bottomrule
\end{tabular}
\end{table}

\subsection{Error threshold}\label{sec8.5}

The parameters for the error thresholds, $\delta(Y) = \delta_0 + bY$, must inherently be application-dependent. For example, soil organic carbon may be assessed for (i) carbon sequestration accounting, where strict absolute accuracy is required, or (ii) soil fertility evaluation, where the required accuracy depends on the concentration range. In this study, parameterisation of $\delta(Y)$ is guided by agronomic soil fertility classes used for Québec \cite{gasser2023rapport} and Canadian soil texture classes \cite{canadian1998canadian}, without reframing the regression task as a classification problem.

For P, K, Al, and pH, fertility classes as defined in Gasser et al.\ \cite{gasser2023rapport} for Québec were used. For K and P, agronomic soil fertility classes typically widen with increasing concentration, reflecting decreasing sensitivity to absolute deviations at higher levels. For instance, K and P are defined as strongly deficient at $<50~\mathrm{mg\,kg^{-1}}$ and $<20~\mathrm{mg\,kg^{-1}}$, respectively. Accordingly, a base tolerance, $\delta_0$, is set to half of these values ($25~\mathrm{mg\,kg^{-1}}$ for K and $10~\mathrm{mg\,kg^{-1}}$ for P). At higher concentrations, class intervals tend to widen, with all values above $225~\mathrm{mg\,kg^{-1}}$ (K) and $125~\mathrm{mg\,kg^{-1}}$ (P) classified as strongly elevated. This increasing tolerance is reflected by setting a relative scaling coefficient of $b = 0.3$. For Al and pH, fertility classes are not systematically widening. Consequently, we set $b = 0$ and define $\delta_0$ as half of the narrowest class interval, yielding $\delta_0 = 250~\mathrm{mg\,kg^{-1}}$ for Al (moderate class: $1{,}100\text{--}1{,}600~\mathrm{mg\,kg^{-1}}$) and $\delta_0 = 0.25$ for pH (low class: 5.5--6). Note that for the latter, we had to rely on classes for pH measured in water.

Soil texture classes in the Canadian system of soil classification are inherently irregular \cite{canadian1998canadian} due to the joint consideration of three particle size fractions. However, clay classes generally widen as clay content increases; for example, values above $600~\mathrm{g\,kg^{-1}}$ are simply classified as heavy clay. We set clay to $\delta_0 = 50~\mathrm{g\,kg^{-1}}$, reflecting approximately half the width of the narrower texture classes at lower clay contents, and set $b = 0.15$ to account for the moderate widening. For the sand fraction, defining $\delta(Y)$ was less straightforward, as class widths are highly irregular, ranging from less than 100 to $400~\mathrm{g\,kg^{-1}}$. We therefore set $\delta_0 = 120~\mathrm{g\,kg^{-1}}$ and $b = 0$, representing an approximate half-width for most, though not all, texture classes.

SOM and TC are closely related, as most soils in Québec contain negligible inorganic carbon (e.g., carbonates). For demonstration purposes, we defined two distinct threshold schemes that reflect different application contexts. For SOM, we relied on Gasser et al.\ \cite{gasser2023rapport}, in which classes widened with increasing SOM. While lower classes may span only $10~\mathrm{g\,kg^{-1}}$, values above $130~\mathrm{g\,kg^{-1}}$ are simply classified as very elevated. Accordingly, we determined $\delta_0 = 5~\mathrm{g\,kg^{-1}}$ with a relative scaling coefficient of $b = 0.3$. In contrast, for TC, we assumed a constant threshold of $10~\mathrm{g\,kg^{-1}}$ with $b = 0$, reflecting applications such as carbon stock estimations that require absolute accuracy.

\subsection{Measurement cost}\label{sec8.6}
As a reference point, we used the per-measurement laboratory costs by IRDA for the economic analysis. However, to preserve confidentiality, we applied a random perturbation to the true cost values, shifting them by up to $\pm 15\%$, and subsequently defined a $\pm 20$--$30\%$ deviation range around these perturbed values for the $y$-axis in Fig.~\ref{fig:Fig4} and Fig.~\ref{fig:FigS6}. The reported measurement costs comprise instrument depreciation, maintenance, labour, and consumables (e.g., chemical reagents), but exclude drying, sieving, and sampling costs.

Explicit measurement costs for VNIRS are rarely reported in the literature. Given the operational simplicity of VNIRS, per-measurement costs are generally assumed to be substantially lower than those of conventional laboratory analyses. An early estimate by Nduwamungu et al.\ \cite{nduwamungu2009opportunities} reported a cost of 4.35~\$CAD per VNIRS measurement and prediction at a laboratory in Québec. However, this estimate predates 2009 and may not reflect current cost structures. Increases in labour and operational costs are likely to raise expenses, whereas advances in instrumentation, automation, and computational efficiency may offset these effects. Nonetheless, the cost estimates for conventional laboratory analyses reported by Nduwamungu et al.\ \cite{nduwamungu2009opportunities} are broadly consistent with those assumed in our analysis. Based on these considerations, we adopted a conservative cost range of 1--10~\$CAD per VNIRS measurement.

\section*{Acknowledgements}

After encountering considerably overoptimistic uncertainty estimates, we thank D.~Holzmüller for the suggestion to rescale the predictive distribution, as implemented in \texttt{s\_rescaler}.

\section*{Author contributions}

Conceptualisation: J.~Schmidinger; R.~Gebbers; V.~Adamchuk.\\
Methodology: All authors.\\
Software: J.~Schmidinger (lead); V.~Barkov.\\
Formal analysis and visualisation: J.~Schmidinger.\\
Data preparation: M.-O.~Gasser (soil data); J.~Schmidinger (VNIRS).\\
Writing—original draft: J.~Schmidinger.\\
Writing—review and editing: All authors.\\
Supervision: V.~Adamchuk; R.~Gebbers.

\section*{Funding}

J.~Schmidinger and V.~Barkov acknowledge support from the Lower Saxony Ministry of Science and Culture (MWK), funded through the \textit{zukunft.niedersachsen} programme of the Volkswagen Foundation, Germany (ZN4072). The Research and Development Institute for the Agri-environment (IRDA) provided in-kind contributions funded through the Québec Ministry of Agriculture, Food and Fisheries (MAPAQ). Spectral equipment was funded through the Natural Sciences and Engineering Research Council of Canada (NSERC) Discovery project on Integrated proximal sensing of soil and crops.

\section*{Data and code availability}

Supporting code can be found on GitHub at \href{https://github.com/JonasSchmidinger/Reject-to-Remeasure}{github.com/JonasSchmidinger/Reject-to-Remeasure}. The data supporting the findings of this study are currently being prepared for open release.

\section*{Extended Data Figures and Tables}

\setcounter{figure}{0}
\renewcommand{\thefigure}{S\arabic{figure}}
\renewcommand{\theHfigure}{S\arabic{figure}}

\setcounter{table}{0}
\renewcommand{\thetable}{S\arabic{table}}
\renewcommand{\theHtable}{S\arabic{table}}

\begin{table}[ht]
\centering
\setlength{\tabcolsep}{9pt}
\renewcommand{\arraystretch}{1.2}
\caption{
\textbf{Predictive performance of the ML models (extended).} Probabilistic performance was evaluated using CRPS (continuous ranked probability score; lower means better predictive distributions), $\mathrm{MPIW}_{0.95}$ (mean width of the 95\% prediction interval; lower indicates sharper distributions), and $\mathrm{QCP}_{0.025}/\mathrm{QCP}_{0.975}$ (quantile coverage probability; better marginal coverage when closer to 2.5\% and 97.5\%, respectively). Point predictions were evaluated with $R^2$ (higher means better fit) and MAE (mean absolute error; lower means less error). The model with the lowest CRPS (in bold) was selected for the extended analysis. See Section~\ref{sec8.3} for details about evaluation metrics.
}
\label{tab:TabS1}
\begin{tabular}{llrrrrrr}
\toprule
Target & Model & CRPS & MPIW$_{0.95}$ & QCP$_{0.025}$ & QCP$_{0.975}$ & $R^2$ & MAE \\
\midrule

\multirow{4}{*}{Sand}
& PLSQR  & 67.46 & 399.09 & 5.67 & 94.33 & 0.80 & 92.56 \\
& QRF    & 66.95 & 497.02 & 6.94 & 96.72 & 0.77 & 93.71 \\
& TabPFN & 48.32 & 321.93 & 4.29 & 96.60 & 0.88 & 67.68 \\
& \textbf{TabICL} & \textbf{47.03} & 315.60 & 3.53 & 97.48 & 0.89 & 65.54 \\

\addlinespace

\multirow{4}{*}{TC}
& PLSQR  & 4.13 & 25.69 & 5.30 & 92.69 & 0.66 & 6.10 \\
& QRF    & 3.84 & 29.12 & 4.04 & 97.98 & 0.70 & 5.41 \\
& \textbf{TabPFN} & \textbf{2.71} & 18.34 & 3.40 & 97.48 & 0.85 & 3.74 \\
& TabICL & 2.86 & 19.44 & 2.65 & 96.72 & 0.84 & 3.92 \\

\addlinespace

\multirow{4}{*}{Al}
& PLSQR  & 101.15 & 568.31 & 8.07 & 93.69 & 0.79 & 139.18 \\
& QRF    & 93.10 & 737.62 & 2.52 & 98.61 & 0.81 & 129.04 \\
& TabPFN & 71.14 & 492.72 & 3.53 & 97.60 & 0.89 & 98.59 \\
& \textbf{TabICL} & \textbf{70.88} & 498.25 & 3.66 & 97.86 & 0.89 & 98.65 \\

\addlinespace

\multirow{4}{*}{pH}
& PLSQR  & 0.13 & 0.74 & 6.43 & 93.69 & 0.73 & 0.18 \\
& QRF    & 0.15 & 1.07 & 2.90 & 97.73 & 0.58 & 0.22 \\
& \textbf{TabPFN} & \textbf{0.11} & 0.76 & 3.40 & 96.72 & 0.78 & 0.15 \\
& TabICL & 0.12 & 0.77 & 4.29 & 97.35 & 0.76 & 0.16 \\

\bottomrule
\end{tabular}
\end{table}

\begin{figure}[t]
    \centering
    \includegraphics[width=0.8\linewidth]{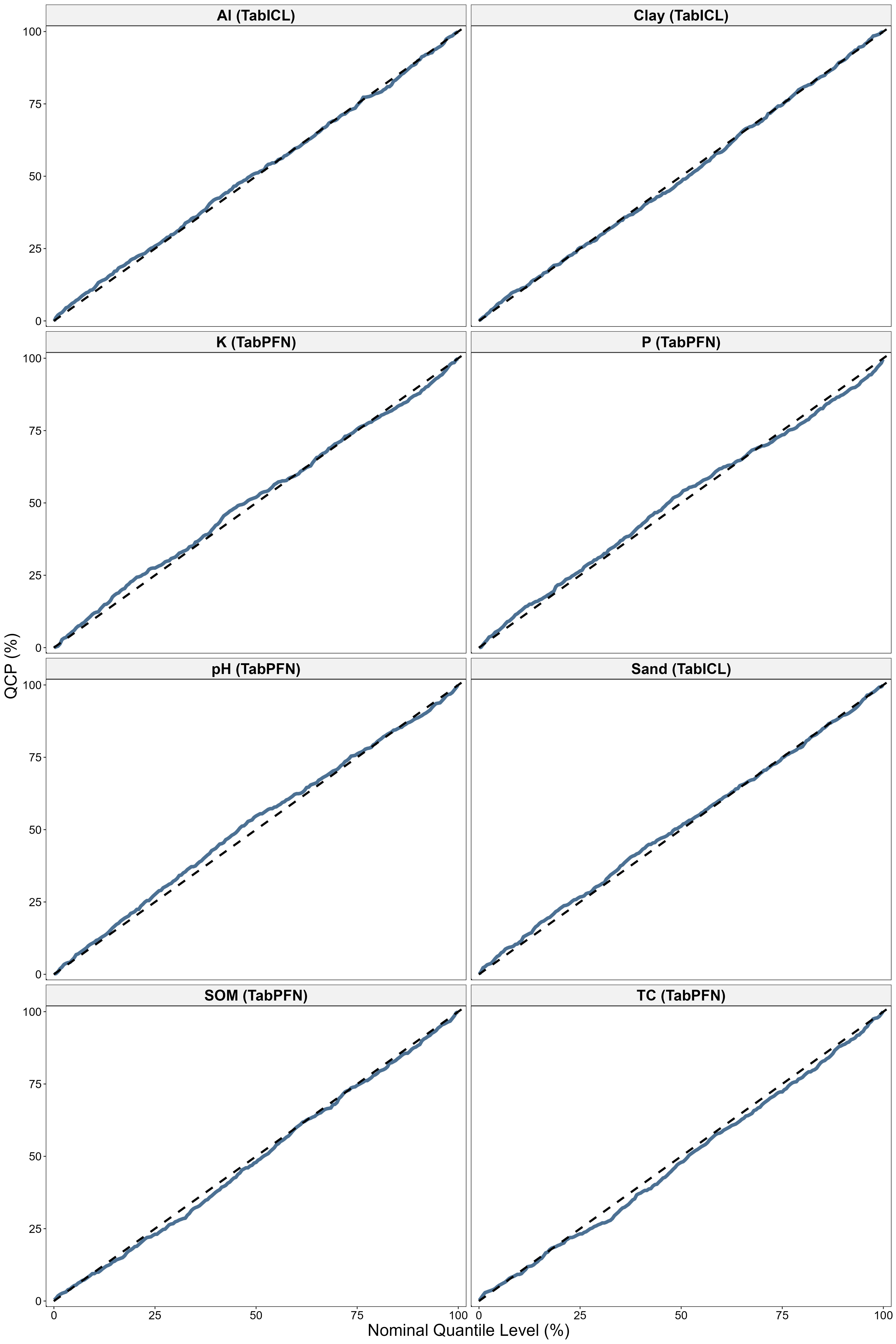}
    \caption{
    \textbf{Reliability plots.} Each subplot shows the empirical QCP against the desired nominal quantile level for a soil property (Clay, SOM, K, P, Sand, TC, Al, and pH). Deviations from the 1:1 line indicate lack of marginal reliability.
    }
    \label{fig:FigS1}
\end{figure}

\begin{figure}[t]
    \centering
    \includegraphics[width=0.8\linewidth]{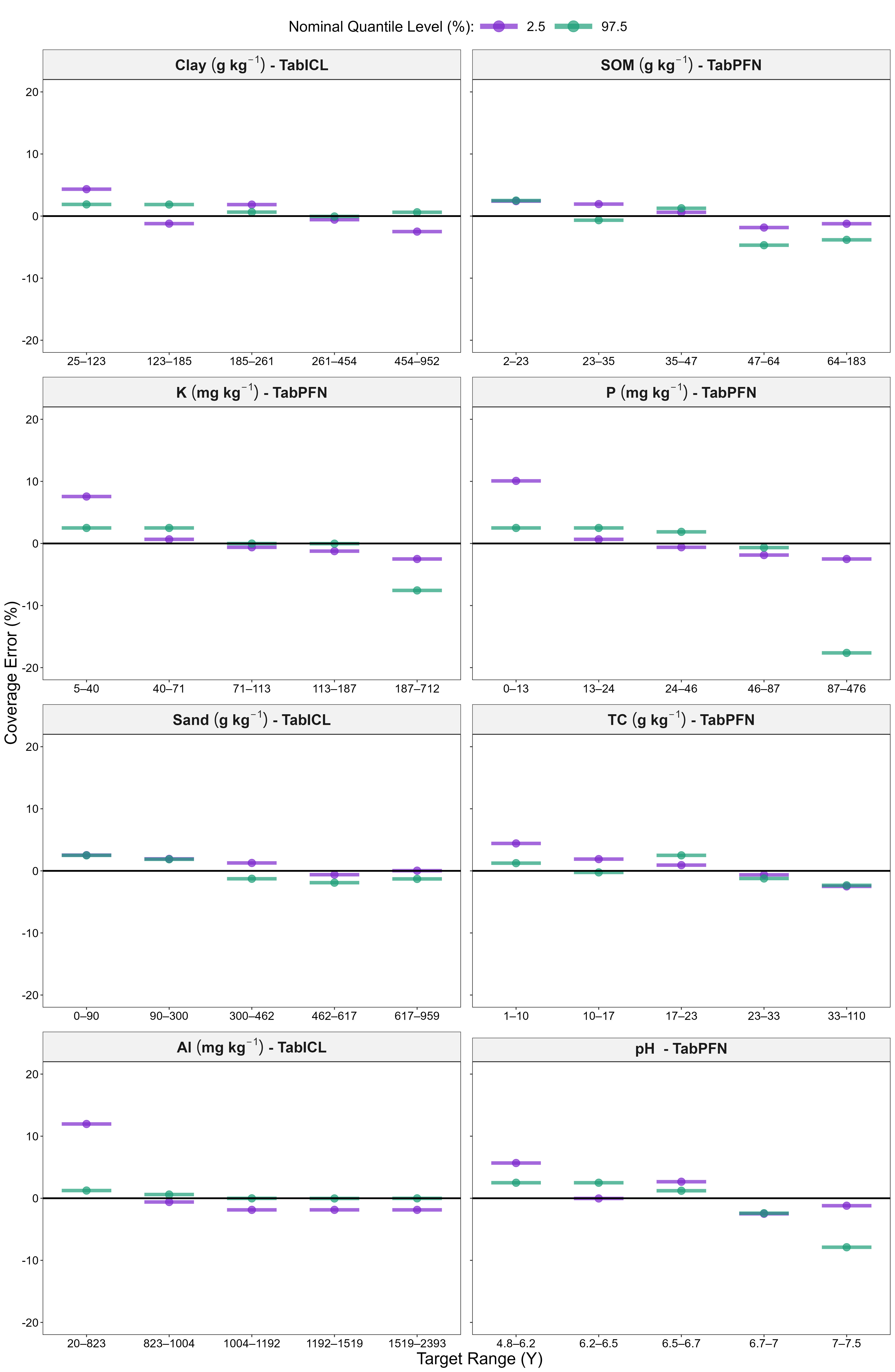}
    \caption{
    \textbf{Conditional coverage error across the range of soil properties.} Each subplot shows the coverage error (deviation between QCP and the nominal quantile level) for different ranges of the target variable, coloured by the 2.5\% and 97.5\% nominal quantile levels. The ranges correspond to quintiles (20\% bins) of the observed soil property values. Different degrees of coverage errors reveal non-ideal conditional reliability.
    }
    \label{fig:FigS2}
\end{figure}

\begin{figure}[t]
    \centering
    \includegraphics[width=\linewidth]{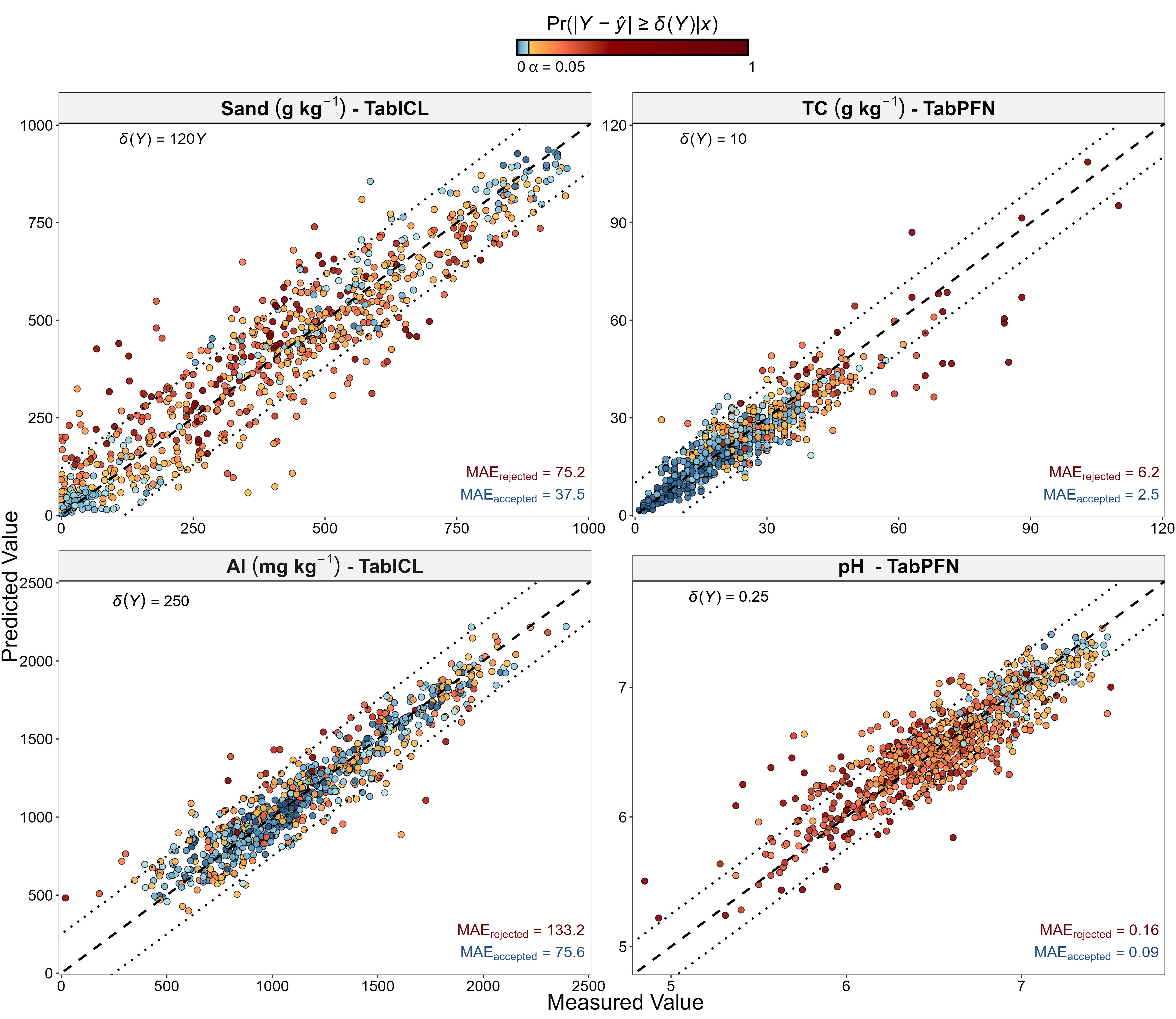}
    \caption{
    \textbf{Predicted versus measured values with uncertainty-based colouring (extended).} Each subplot shows one soil property (Sand, TC, Al, and pH), with predicted values plotted against measured values. The dashed line represents the 1:1 predicted–measured relationship, and the dotted lines indicate the property-specific error threshold $\delta(Y)$ (see Section~\ref{sec8.5}). Colours indicate the model’s estimated probability that the absolute prediction error exceeds $\delta(Y)$. Predictions with probabilities below $\alpha = 0.05$ are accepted by the rejector (blue gradient), whereas higher probabilities correspond to rejected predictions (orange–red gradient). $\mathrm{MAE}_{\text{accepted}}$ and $\mathrm{MAE}_{\text{rejected}}$ denote the mean absolute error of the accepted and rejected predictions, respectively.
    }
    \label{fig:FigS3}
\end{figure}

\begin{figure}[t]
    \centering
    \includegraphics[width=\linewidth]{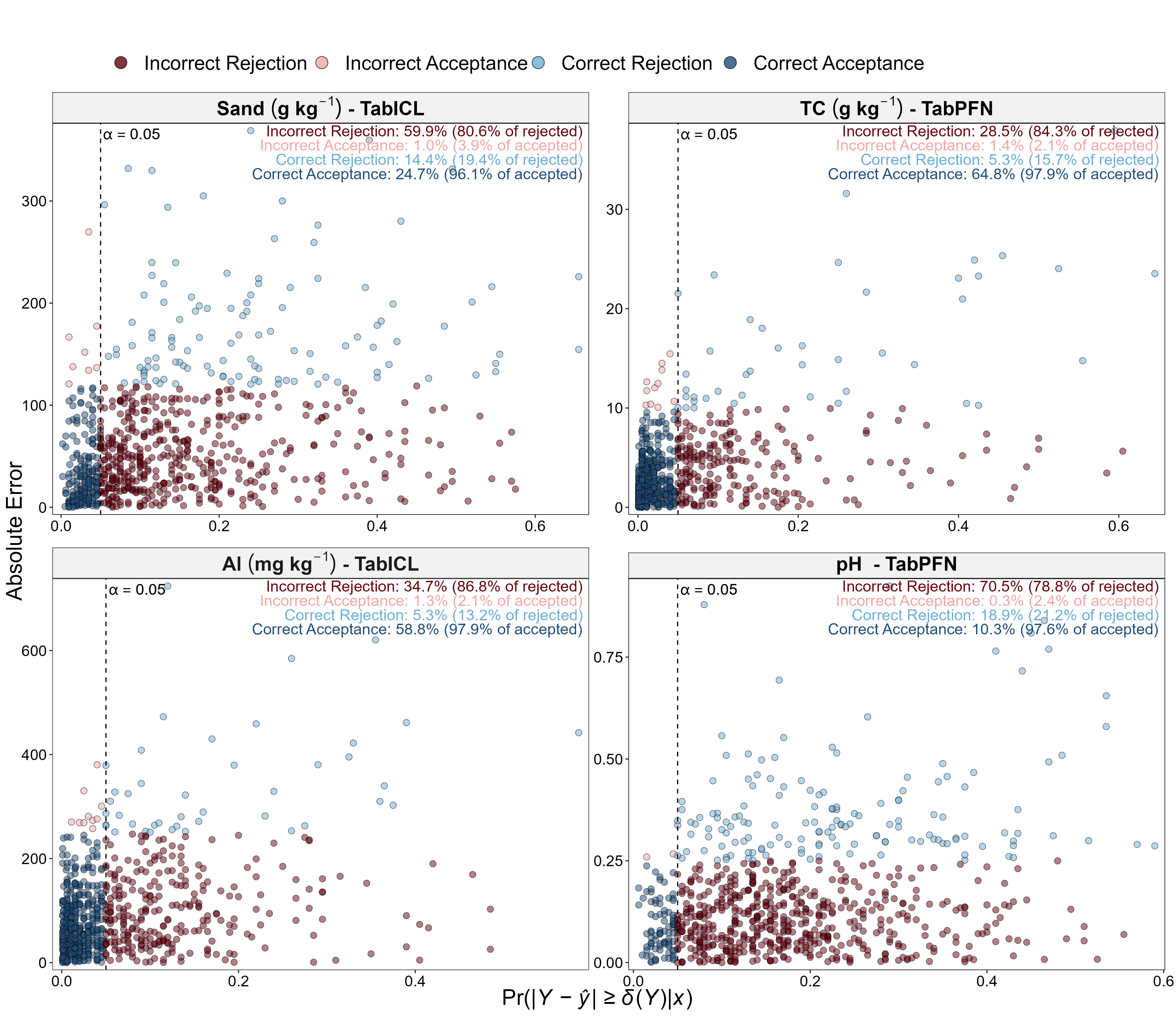}
    \caption{
    \textbf{Rejector decision outcomes (extended).} Each subplot shows one soil property (Sand, TC, Al, and pH), with the predicted probability that the absolute error exceeds the threshold $\delta(Y)$, i.e., $\Pr\!\left(\lvert Y - \hat{y}(x) \rvert \geq \delta(Y) \mid x \right)$, plotted against the apparent absolute error. The vertical dashed line indicates the rejector decision boundary corresponding to the risk tolerance $\alpha = 0.05$; predictions to the left are accepted, whereas predictions to the right are rejected. Points are coloured according to the resulting decision outcome. Percentages denote the fraction of samples in each category. Values outside parentheses refer to the percentage relative to all samples, whereas values in parentheses refer to the percentage relative to accepted or rejected predictions.
    }
    \label{fig:FigS4}
\end{figure}

\begin{figure}[t]
    \centering
    \includegraphics[width=0.8\linewidth]{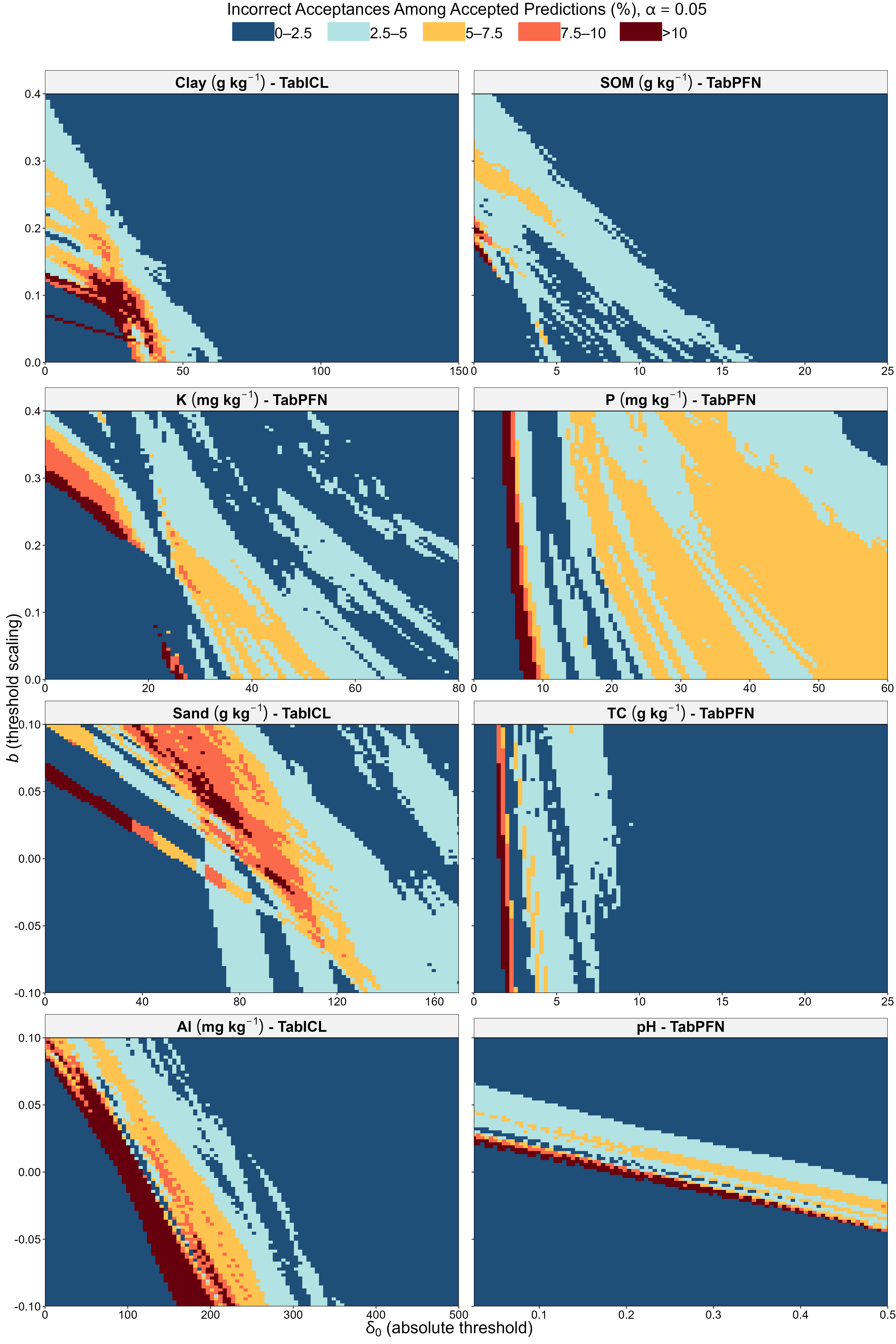}
    \caption{
    \textbf{Incorrect acceptance rate relative to the error threshold parameters (extended).} Each subplot shows the rate of incorrect acceptances among accepted predictions for a soil property (Sand, TC, Al, and pH), as a function of the parameters in the error threshold $\delta(Y) = \delta_0 + bY$, where $\delta_0$ is the absolute threshold component and $b$ the scaling coefficient. Depending on the parameterisation, the $\alpha = 0.05$ quality constraint is satisfied (blue) but may be violated in certain cases (orange to red) due to suboptimal conditional reliability.
    }
    \label{fig:FigS5}
\end{figure}

\begin{figure}[t]
    \centering
    \includegraphics[width=\linewidth]{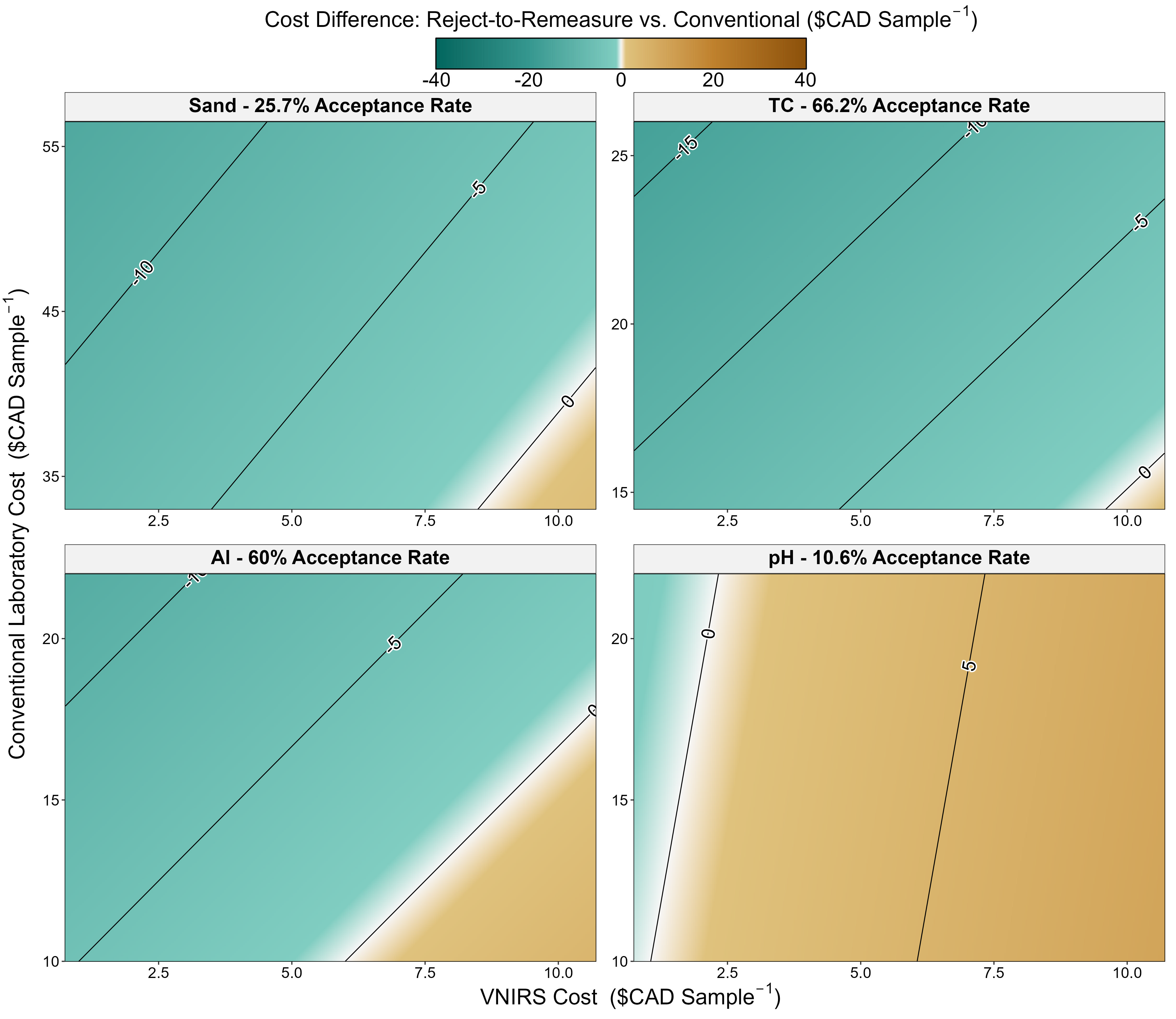}
    \caption{
    \textbf{Cost difference between VNIRS reject-to-remeasure and conventional laboratory testing (extended).} Each subplot shows the cost difference in the soil property analysis (Sand, TC, Al, and pH) when using the VNIRS-based reject-to-remeasure strategy compared to pure conventional laboratory analysis. Plausible ranges of measurement costs for conventional analysis and VNIRS were evaluated (see Section~\ref{sec8.6}). Negative values (blue–green) indicate cost savings with reject-to-remeasure as compared to the conventional analysis, whereas positive values (brown) indicate increased measurement costs. Contour lines denote equal levels of cost savings. The plot assumes the acceptance rates obtained in the previous rejector evaluation (see Section~\ref{sec8.5}).
    }
    \label{fig:FigS6}
\end{figure}

\begin{figure}[t]
    \centering
    \includegraphics[width=\linewidth]{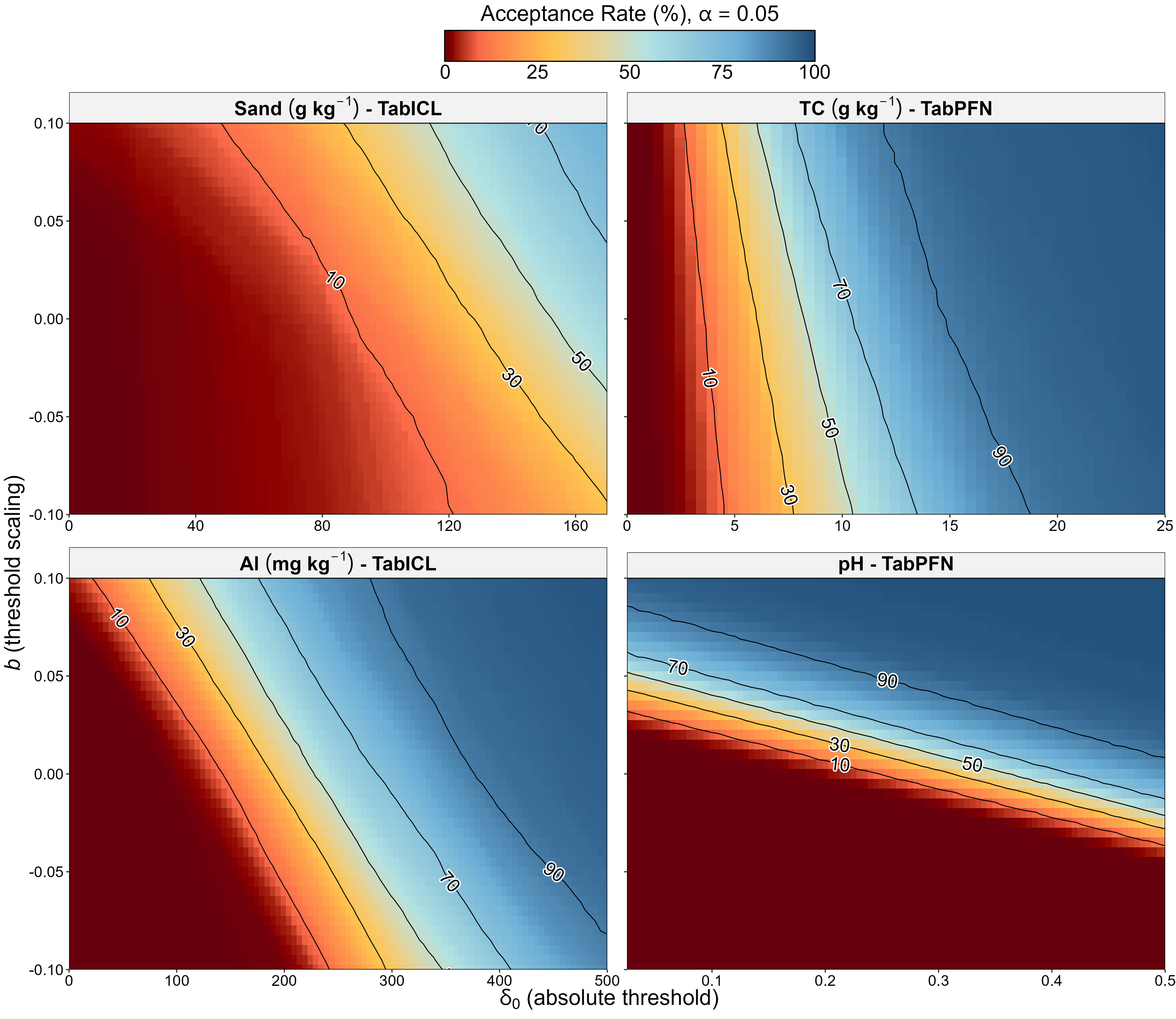}
    \caption{
    \textbf{Acceptance rates relative to the error threshold parameters (extended).} Each subplot shows the acceptance rate for a soil property (Sand, TC, Al, and pH) as a function of the parameters in the error threshold $\delta(Y) = \delta_0 + bY$, where $\delta_0$ is the absolute threshold component and $b$ the scaling coefficient. With higher error tolerances, acceptance rates increase (blue), whereas overly strict quality constraints reduce the acceptance rate to near zero (red). Contour lines denote equal acceptance rates.
    } 
    \label{fig:FigS7}
\end{figure}

\end{document}